\documentclass{article}

\usepackage[final]{neurips_2022}

\usepackage[utf8]{inputenc} %
\usepackage[T1]{fontenc}    %
\usepackage{url}            %
\usepackage{booktabs}       %
\usepackage{amsfonts}       %
\usepackage{nicefrac}       %
\usepackage{microtype}      %
\usepackage{xcolor}         %
\usepackage{microtype}
\usepackage{graphicx}
\usepackage{amsmath}
\usepackage{multirow}
\usepackage{subcaption}
\usepackage{algorithm}
\usepackage{algpseudocode}
\usepackage[pagebackref=true]{hyperref} 
\hypersetup{colorlinks, citecolor=purple, linkcolor=purple}
\usepackage{wrapfig}
\usepackage{nameref}
\usepackage{pifont}
\usepackage[most]{tcolorbox}

\everypar{\looseness=-1}
\renewcommand*\backref[1]{\ifx#1\relax \else (Cited on #1) \fi}

\usepackage{amsmath,amsfonts,bm}

\def\eqref#1{equation~\ref{#1}}

\def\1{\bm{1}}

\def\eps{{\epsilon}}

\def\vw{{\bm{w}}}
\def\vx{{\bm{x}}}
\def\vy{{\bm{y}}}
\def\vz{{\bm{z}}}

\DeclareMathAlphabet{\mathsfit}{\encodingdefault}{\sfdefault}{m}{sl}
\SetMathAlphabet{\mathsfit}{bold}{\encodingdefault}{\sfdefault}{bx}{n}

\newif\ifcomments

\commentsfalse

\ifcomments
  \newcommand{\colornote}[3]{{\color{#1}\bf{#2: #3}\normalfont}}
\else
  \newcommand{\colornote}[3]{}
\fi

\usepackage[capitalize]{cleveref}
\crefname{equation}{Eq.}{Eq.}
\crefname{figure}{Fig.}{Fig.}
\crefname{table}{Tab.}{Tab.~}
\crefname{section}{Sec.}{Sec.~}
\crefname{algorithm}{Alg.}{Alg.~}
\crefname{thm}{Theorem}{Theorem~}
\crefname{lemma}{Lemma}{Lemma~}
\crefname{appendix}{Appendix}{Appendix~}

\newcommand{\reals}{\mathbb{R}}
\newcommand{\pp}[2]{\frac{\partial #1}{\partial #2}}
\newcommand{\dd}[2]{\frac{d#1}{d#2}}

\def\ie{\textit{i.e.,~}}

\algnewcommand{\LeftComment}[1]{\Statex \(\triangleright\) #1}

\newcommand{\cmark}{\ding{51}}%
\newcommand{\xmark}{\ding{55}}%
\newcommand{\efficacyhigh}{\textcolor{green}{\cmark\cmark}}
\newcommand{\efficacymed}{\textcolor{orange}{\cmark}}
\newcommand{\efficacylow}{\textcolor{red}{\xmark}}

\title{Path Independent Equilibrium Models Can Better Exploit Test-Time Computation}

\author{
Cem Anil$^{1}$\thanks{Equal contribution. Correspondence to anilcem@cs.toronto.edu and apokle@cs.cmu.edu.} \quad 
Ashwini Pokle$^{2*}$ \quad
Kaiqu Liang$^{3*}$ \quad
Johannes Treutlein$^{1,4}$ \quad \\
\textbf{Yuhuai Wu}$^5$ \quad
\textbf{Shaojie Bai}$^{2}$ \quad
\textbf{Zico Kolter}$^{2,6}$ \quad
\textbf{Roger Grosse}$^{1}$\\
$^1$University of Toronto and Vector Institute \quad $^2$Carnegie Mellon University 
\quad $^3$Princeton University \\ $^4$University of California, Berkeley
$^5$Stanford University and Google Research \quad $^6$Bosch Center for AI \\
\texttt{\{anilcem, rgrosse\}@cs.toronto.edu} \quad \texttt{kl2471@princeton.edu}\\
\texttt{\{apokle, shaojieb, zkolter\}@cs.cmu.edu}\\
\texttt{yuhuai@google.com} \quad \texttt{johannestreutlein@berkeley.edu}
}

\begin{document}

\maketitle

\begin{abstract}
Designing networks capable of attaining better performance with an increased inference budget is important to facilitate generalization to harder problem instances. Recent efforts have shown promising results in this direction by making use of depth-wise recurrent networks. We show that a broad class of architectures named \textit{equilibrium models}  display strong upwards generalization, and find that stronger performance on harder examples (which require more iterations of inference to get correct) strongly correlates with the \emph{path independence} of the system---its tendency to converge to the same steady-state behaviour regardless of initialization, given enough computation.
Experimental interventions made to promote path independence result in improved generalization on harder problem instances, while those that penalize it degrade this ability. Path independence analyses are also useful on a per-example basis: for equilibrium models that have good in-distribution performance, path independence on out-of-distribution samples strongly correlates with accuracy. Our results help explain why equilibrium models are capable of strong upwards generalization and motivates future work that harnesses path independence as a general modelling principle to facilitate scalable test-time usage. 
\end{abstract}

\section{Introduction} \label{sec:introduction}

One of the main challenges limiting the practical applicability of modern deep learning systems is the ability to generalize outside the training distribution \citep{koh2021wilds}. One particularly important type of out-of-distribution (OOD) generalization is \emph{upwards generalization}, or the ability to generalize to more difficult problem instances than those encountered at training time \citep{selsam2018learning, bansal2022end, schwarzschild2021can, nye2021show}. Often, good performance on more difficult instances will require a larger amount of test-time computation, so a natural question arises: how can we design neural net architectures which can reliably exploit additional test-time computation to achieve better accuracy?

Equilibrium models, a broad class of architectures whose outputs are the fixed points of learned dynamical systems, are particularly suited to meet this challenge. Closely related to weight-tied recurrent models -- networks that apply the same fixed neural network module repeatedly to hidden-layer activations -- equilibrium models are capable of adapting their compute budget based on the input they are given. Under what conditions, if any, can this input-dependent ability to scale-up test-time compute actually lead to upwards generalization?

We argue that a key determiner of whether a learned model can exploit additional test-time computation is whether the dynamical system corresponding to the model is \emph{path independent}; that is, whether the learned model's hidden layer activations converge to the same asymptotic behaviour (i.e. fixed point or limit cycle), regardless of the initialization of the system. For example, a simple integrator $x_{t+1} = x_t + 1$ is clearly \emph{not} path independent, as its final state depends on the initial state $x_0$ and the number of iterations run; conversely, the system $x_{t+1} = (x_t+1)/2$ \emph{is} path independent, as it will converge to the solution $x_T = 1$ as $T \rightarrow \infty$ regardless of the initial condition of $x_0$. Path independence is closely related to the concept of \textit{global stability} from control theory (see Section \ref{sec:related} for more).

Intuitively, path independent systems can more easily take advantage of additional test-time iterations than path dependent ones. For instance, gradient descent applied to a convex objective is path independent, and correspondingly when confronted with a more ill-conditioned problem instance, one can compensate by increasing the number of iterations. Conversely, a weather simulation is path dependent, and extending the simulation won't yield more accurate predictions of a given day's weather. Based on this intuition, we hypothesize that path independence of a learned model is a key determiner of whether it can take advantage of an increased test-time iteration budget when generalizing to harder problem instances:

\begin{figure*}[!]
\centering
\includegraphics[width=0.77\textwidth]{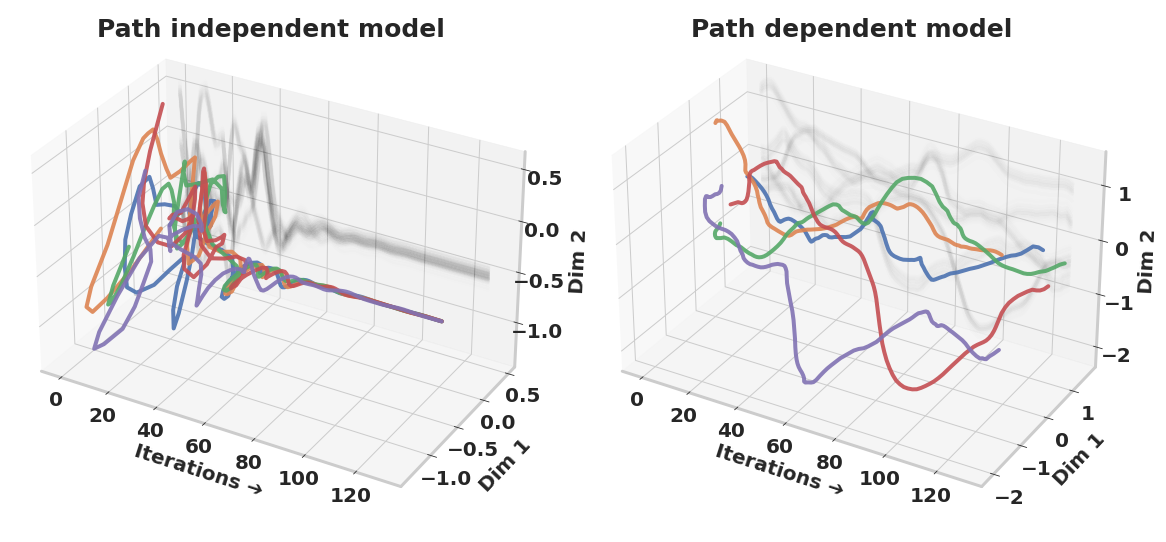}

\caption{Trajectories of path independent models converge to the same hidden state for a given input, regardless of initialization, whereas the trajectories of path dependent models depend on initialization. Here, we display five trajectories with different initializations obtained from a path independent (left) and path dependent model (right) on the prefix-sum task, projected onto two random directions.}

\label{fig:trajectory_plot}
\vspace{-4mm}
\end{figure*}

\begin{quote}
\emph{Path Independence Hypothesis:} Models which successfully fit the training distribution with a path independent function are better able to exploit more test-time iterations to achieve higher accuracy, compared to those which fit the training distribution with a path-dependent function. 
\end{quote}

We first introduce a metric for measuring the path independence of a learned model, the \emph{Asymptotic Alignment (AA)} score. On a wide range of tasks including sequence prediction, visual reasoning, image classification, continuous optimization and graph regression, we show that a model's AA score is strongly correlated with its performance when far more iterations are used at test time than at training time. In general, we find that path independent models increase their performance monotonically with the number of test-time iterations, whereas path dependent models degrade when the number of test time iterations exceeds those at training time. We find that input injection and implicit updates improve both the path independence and the accuracy. Furthermore, we perform an experimental manipulation by introducing regularizers which directly promote or punish path independence. We find that these manipulations, while increasing or decreasing the AA score, also have the corresponding effect on accuracy.
\vspace{-3mm}
\section{Background}
\label{sec:background}

\paragraph{Equilibrium Models} Equilibrium models treat computing internal representations as a fixed-point finding problem.~\citep{mcclelland1989explorations, liao2018reviving, bai2019deep} Concretely, letting $\vx \in \reals^{n_x}$ and $f_{\vw}: \reals^{n_x \times n_z} \mapsto \reals^{n_z}$ stand for an input and the equilibrium model function (or ``cell") parametrized by $\vw \in \reals^{n_w}$ respectively, equilibrium models aim to solve for the fixed point $\vz^* \in \reals^{n_z}$ that satisfies $\vz^* = f_{\vw}(\vx, \vz^*)$. The cell $f_{\vw}$ might represent anything from a fully connected layer to a transformer block~\citep{vaswani2017attention}. We emphasize that $f_{\vw}$ directly depends on the input $\vx$; following existing literature, we refer to this as \textit{input injection}. The outcome of the fixed point finding process might depend on the initial fixed point guess. To make this dependency explicit, we define the function $\mathrm{FIX}: \reals^{n_x \times n_z} \mapsto \reals^{n_z}$ that maps an input $\vx$ and an initial guess for the fixed point $\vz_0$ to an output that satisfies the fixed point equation $\mathrm{FIX}_{f_w}(\vx, \vz_0) := f_{\vw}(\vx, \mathrm{FIX}_{f_w}(\vx, \vz_0))$. The behaviour of $\mathrm{FIX}$ depends on the solver that's used to find fixed points. 

The most straightforward approach to solve for fixed-points is the \textit{fixed point iteration} method, which recursively applies the function $f_{\vw}$ on the internal representations $\vz$ (i.e. $\vz_{t+1} = f_{\vw}(\vx, \vz_{t})$). If certain conditions are satisfied (such as the fixed iterations being globally contractive -- more general conditions are discussed below), this procedure converges\footnote{Divergence is also a possible outcome, rendering the output of equilibrium models unusable.} to a fixed point: $\vz^* = f_{\vw}(\vx, \vz^*)$. As solving for fixed points exactly is expensive, fixed point iterations are often terminated after a fixed number of steps or when the norm of the difference between subsequent iterates falls below a pre-selected threshold. The model weights can be updated using gradients computed via backpropagating through the full forward computational graph. 

If one commits to using fixed point iterations as the root solver, then the output $\vz^*$ of equilibrium models can be interpreted as the infinite-depth limit of an input-injected, weight-tied model $f^{\infty}_{\vw}(\vx, \vz^*) = \lim_{n \to \infty} f^{(n)}_{\vw}(\vx, \vz_0) $ where the notation $f^{(n)}$ stands for $n$ repeated applications of $f$ on its own output, and $\vz_0$ stands for the fixed point initialization. 

\textbf{Implicit Training of Equilibrium Models} Different training algorithms for equilibrium models can be derived by considering their implicit nature. \citet{bai2019deep} solve for fixed points explicitly using black-box root finders, such as Broyden's method \citep{broyden1965class} or Anderson acceleration \citep{anderson1965iterative}. In order to avoid explicitly differentiating through the root-finding procedure, they utilize implicit differentiation to compute gradients.\footnote{\citet{bai2019deep} use the term \textit{Deep Equilibrium Models}~(DEQ) to refer to implicitly trained equilibrium models. To keep things more general, we categorize ``explicitly trained" networks (i.e. with fixed point iterations and backpropagation) under the umbrella of equilibrium models as well.} Concretely, letting $g_{\vw}(\vx, \vz) = f_{\vw}(\vx, \vz) - \vz$ for a fixed point $\vz^*$, the Jacobian of $\vz^*$ with respect to the equilibrium model weights can be given by:
\begin{equation}
    \dd{\vz^*}{\vw} = -\left(\pp{g_{\vw}(\vx, \vz^*)}{\vz^*}\right)^{-1}\pp{f_{\vw}(\vx, \vz^*)}{\vw}
    \label{eq:deq-backward-pass}
\end{equation}

Inverting a Jacobian matrix can become computationally expensive. Recent works \citep{geng2021attention, fung2021fixed} have shown that the inverse-Jacobian term in \cref{eq:deq-backward-pass} can be replaced with an identity matrix i.e. Jacobian-free or an approximate inverse-Jacobian \citep{geng2021training} without affecting the final performance. This approximation makes the backward pass inexpensive and lightweight. Equilibrium models have been shown to achieve state-of-the-art performance on various tasks including language modelling \citep{bai2019deep}, image recognition,  semantic segmentation \citep{bai2020multiscale}, object detection \citep{wang2020implicit}, and graph modeling \citep{gu2020implicit, Liu2021EIGNNEI}.

\textbf{Equilibrium Models vs.~Depthwise Recurrent Models} Both equilibrium models and input-injected depthwise recurrent (i.e. weight-tied, fixed-depth) networks leverage weight-tying \ie they apply the same transformation at each layer, $f_{\omega}^{[i]} = f_{\omega} \; \forall i$. The two models differ in the ultimate aim of the forward pass: while depthwise recurrent models compute a (weight-tied) fixed depth computation (which may or may not approach a fixed point), the stated \emph{goal} of equilibrium models is explicitly to find a fixed point. Weight-tied fixed depth networks by definition require backpropagation through an explicit stack of layers. Equilibrium models, however, directly solve for fixed points using (potentially black-box) solvers during the forward pass and may be trained using implicit differentiation. 

\textbf{Convergence}
As alluded above, in order to guarantee convergence to a unique fixed point, it suffices for the cell of the equilibrium model to be \textit{contractive} over its input domain\footnote{This is known as the Banach fixed-point theorem.} (i.e. the singular values of its Jacobian all lie below $1$). Previous work has leveraged Lipschitz constrained cells to ensure contractivity \citep{revay2020lipschitz}. Other approaches for ensuring global convergence exist: the monotone equilibrium model architecture guarantees global convergence by utilizing an equilibrium model parametrization that bears similarities to solutions to a particular form of monotone operator splitting problem~\citep{winston2020monotone}. Unrestricted equilibrium models aren't constrained enough to guarantee convergence: they can easily express globally divergent vector fields that prohibit the existence of fixed points. It is, therefore, interesting that they can (and often) learn path independent solutions. Also note that the ``infinite-depth weight tied network" interpretation of equilibrium models is less general than the implicit formulation presented above, as the latter admits unstable fixed points as well. 

\textbf{Terminology and Abbreviations} We use the term ``equilibrium models'' to refer to the general class of networks that explicitly solve for a fixed point in the forward pass. We use the term `solver' to refer to the use of black-box root finders like Anderson acceleration to find fixed points of an implicitly trained equilibrium model. These networks can use implicit gradients computed via implicit function theorem (IFT), Jacobian-free backward pass, or with an approximation of inverse-Jacobian. The term `unroll' refers to equilibrium models that use regular fixed-point iterations to compute the equilibrium point. We use the abbreviation `bp' to refer to backpropagation gradients, and `inj' to refer to input injection. The term `progressive net' refers to the deep thinking networks trained with progressive training as proposed by \citet{bansal2022end}. We use `PI' and `non-PI' to refer to path independent and path dependent networks, respectively.
\vspace{-3mm}
\section{Upwards Generalization with Equilibrium Models}
\label{sec:algo_ood_eq}
\vspace{-1mm}

In this section, we establish that equilibrium models are capable of strong upwards generalization. To study the effects of test time computation, it is useful to consider tasks with an explicit difficulty parameter, so that the learned models can be tested on more difficult instances which require a large number of iterations to solve correctly.  We focus on multiple algorithmic generalization tasks: \textbf{prefix sum} and \textbf{mazes} by \citet{schwarzschild2021datasets, schwarzschild2021can}, \textbf{blurry MNIST}, \textbf{matrix inversion} and \textbf{edge copy} by \citet{du2022learning}. Taken together, these tasks cover a wide range of problems from different domains, namely sequence prediction, visual reasoning, image classification, continuous optimization and graph regression. To maintain clarity and focus, we run our detailed analysis on the prefix sum and mazes tasks, and provide complementary results for the remaining tasks in the Supplementary Material (SM).

\textbf{Tasks} \textbf{Prefix-sum} is a sequence-to-sequence task whereby the network is given a sequence of 0-1 bits, and is trained to output, for each bit, the parity of all of the bits received since the beginning of the sequence until the current bit. We train on 10,000 unique 32-bit binary strings, and report results on binary strings of other lengths. The \textbf{mazes} task is also an image-to-image task, where the input is a three-channel RGB image. The `start' and `finish' positions are marked by a red and a green square respectively; walls are marked in black. The output is the optimal path in the maze that connects these two points without passing through the walls. We train on 50,000 small mazes of size $9\times9$, and report upward generalization results on larger mazes. Instances of each of these problems, as well as additional image classification and continuous optimization results can be found in the supplementary material. \textbf{Blurry MNIST}~\citep{liangout} is a robustness-to-corruption task: one has to learn to do MNIST classification from lightly blurred images and generalize zero-shot to highly blurred ones. In the\textbf{ matrix inversion task}~\citep{du2022learning}, the goal is to learn to invert $10\times10$ matrices in a way that generalizes to matrices that have worse condition number than those observed during training. \textbf{Edge copy}~\citep{du2022learning} is a simple graph regression tasks that requires learning to output the input edge features, in a way that generalizes to larger graph sizes. Note that the training and test data in the latter two tasks are generated with noise added on-the-fly, as done by \citet{du2022learning}. 

\begin{figure}[!t]
  \centering
  \begin{minipage}[b]{0.49\textwidth}
    \centering
    \includegraphics[width=0.99\textwidth]{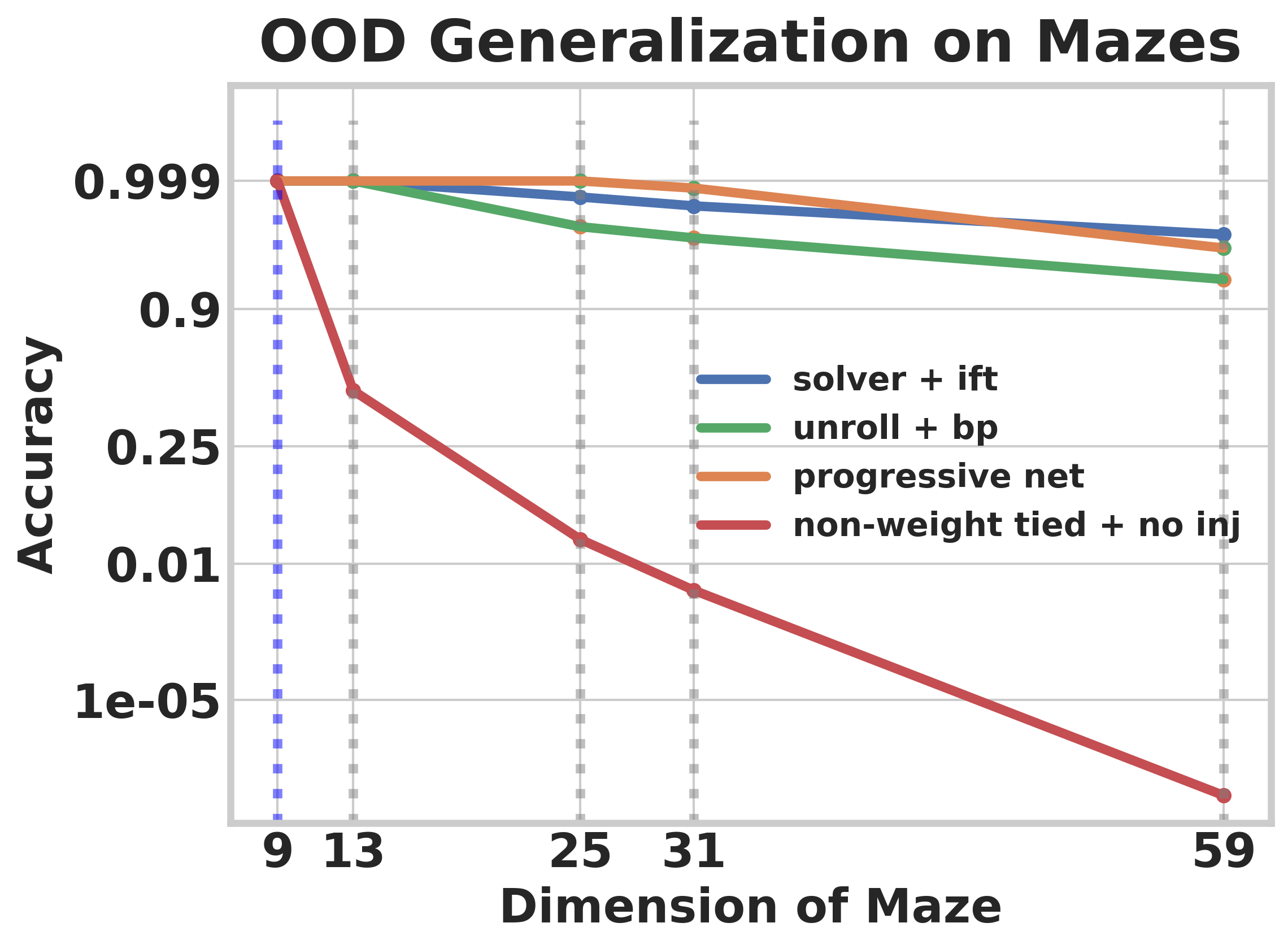}
    \subcaption[]{}\label{fig:ood_generalization_mazes}
  \end{minipage}
  \hfill
  \begin{minipage}[b]{0.49\textwidth}
    \centering
    \includegraphics[width=0.99\textwidth]{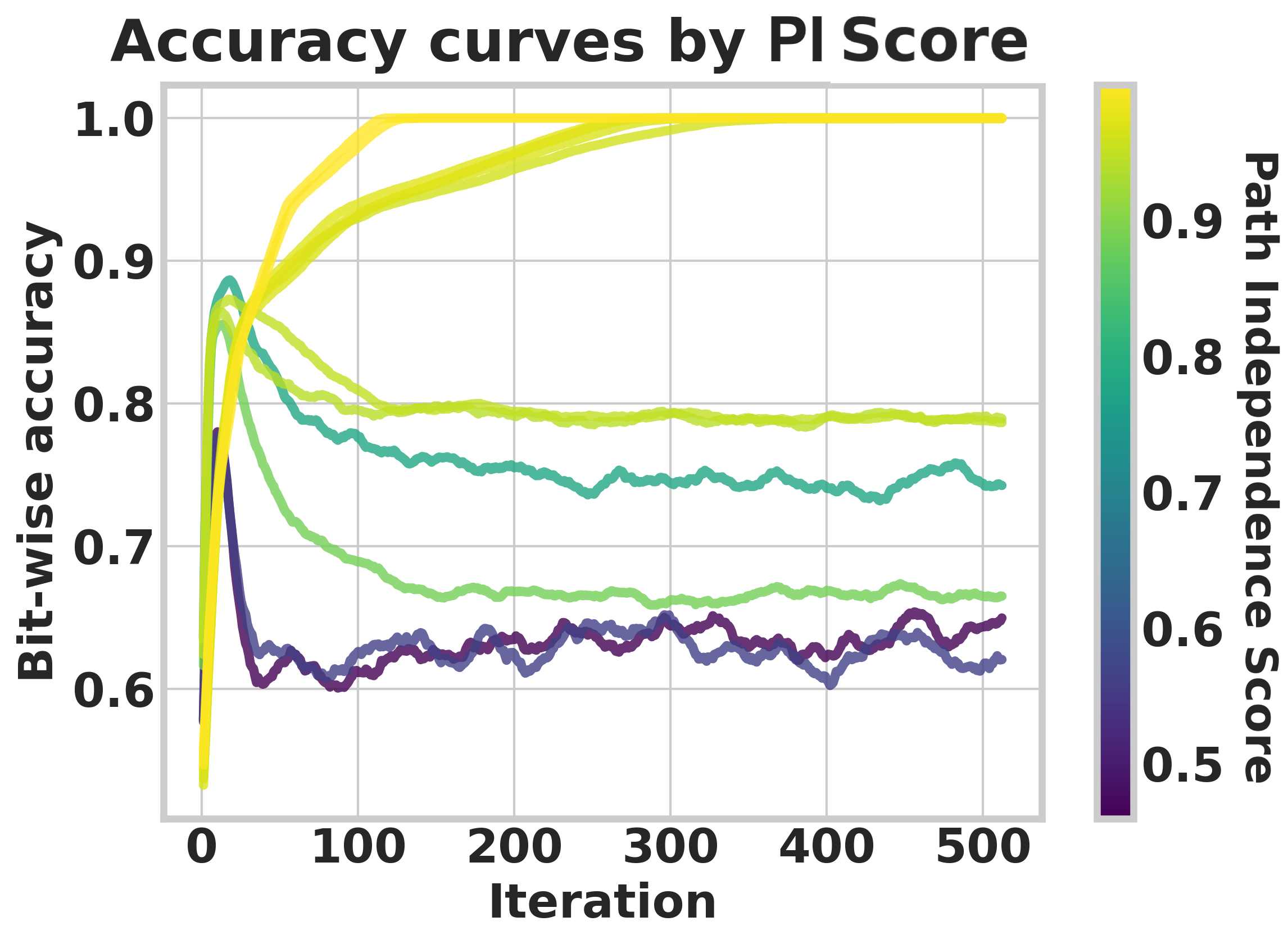}
    \subcaption[]{}\label{fig:pi-vs-nonpi-accuracy}
  \end{minipage}
  \caption{(\textbf{left}) Strong upward generalization on mazes by PI models. Models were trained on $9\times9$ sized mazes and tested for upward generalization on larger mazes. y-axis uses probit transformation. (\textbf{right}) PI models are better able to make use of additional test-time computation. We trained models with varying number of training-time iterations, learning rate and weight norm application. Bit-wise accuracies are evaluated and averaged over different string-lengths.}
\end{figure}

\textbf{Strong Upward Generalization}
\cref{fig:ood_generalization_mazes} shows that equilibrium models demonstrate very strong upward generalization performance compared to non weight-tied fixed-depth models. Moreover, \cref{fig:pi-vs-nonpi-accuracy} shows that increasing inference depth consistently improves performance---especially on harder problem instances.

\vspace{-3mm}
\section{Path Independence}
\vspace{-1mm}

Having intuitively motivated the idea of path independence in \cref{sec:introduction}, we now define it formally: we say that the computation performed by a recurrent operator computing function $f_{\vw}$ on an input $\vx$ is path independent if it converges to the same limiting behavior regardless of the current state $\vz_t$. As a special case, if the computation is convergent, this property is equivalent to the existence of a unique fixed point $\vz^*$ such that $f^{\infty}_{\vw}(\vx, \vz_0) = \vz^*$ for any $\vz_0$. However, our definition allows for other behaviors such as limit cycles (see \cref{sec:disambiguate-convergence-pi}).

Some architectures guarantee the path independence property (see \cref{sec:background}). However, most common DEQ architectures---and the ones we use throughout this paper---have the expressive power to learn multiple fixed points per input. Since it is unclear whether architectures enforcing the contraction property lose expressiveness \citep{Bai2021StabilizingEM}, we focus our investigation on unrestricted architectures. 

PI networks represent a different model for computation than standard feed-forward networks: instead of learning an entire computational graph to map inputs to outputs, they only have to \textit{learn where to stop}. We dedicate the rest of the paper on exploring the \textit{Path Independence Hypothesis}---the idea that models which successfully fit the training distribution with a path independent function are better able to exploit more test-time iterations to achieve higher accuracy, compared to those which fit the training distribution with a path-dependent function.

Before establishing a connection between path independence and out-of-distribution generalization, we first describe two architectural components that are \textit{necessary conditions} for achieving path independence. Afterward, we describe a metric to quantify how path-independent a trained network is. 

\vspace{-2mm}
\subsection{Architectural Components Necessary for Path Independence} 
\label{sec:pi_necessary}
Past work has observed that \emph{weight tying} and \emph{input injection} are both crucial for upwards generalization \citep{bansal2022end}. We observe that both architectural components are also necessary for a learned model to be PI.\footnote{Our definition also admits non-input-injected models to be path independent if they're representing constant functions (i.e. input independent). We don't consider such cases in our analyses.} Without weight tying, the network is constrained to have a fixed forward depth, so it is meaningless to talk about the limiting behavior in large depth. Input injection ensures that the equilibrium point depends on the input despite having an ``infinite depth". Without input injection, a PI network would necessarily forget the input; hence, any model which successfully fits the training distribution must be path dependent. 

Interestingly, both architectural motifs are also key components of deep equilibrium models~\citep{bai2019deep}; in that work, the motivation was to enable efficient gradient estimation via the implicit function theorem (IFT) --- a concept closely related to path independence, since the premise of the IFT gradient estimator is that only the final hidden state matters, not the path taken to get there. It is striking that two separate lines of work would converge on the same architectural motifs, one motivated by generalization and the other by a variant of path independence. 

Reproducing the results of \cite{bansal2022end}, in \cref{fig:ood_generalization_mazes} we show upward generalization performance using both equilibrium models and progressive nets~\citep{bansal2022end} -- and the lack thereof using non-input-injected networks. For the remainder of this paper, we focus on architectures with both input injection and weight tying.

\vspace{-2mm}
\subsection{Quantifying Path Independence}
\vspace{-1mm}

\begin{figure}[!t]
    \centering
    \begin{minipage}[b]{0.46\textwidth}
        \centering
        \begin{algorithm}[H]
            \centering
            \caption{Asymptotic Alignment Score} \label{alg:fpa-score}
            \begin{algorithmic}
            \Require A batched input $\begin{bmatrix}\textcolor{teal}{\vx_1}\\ \textcolor{magenta}{\vx_2} \end{bmatrix}$, an operator $f_\vw$\\
            \textbf{Initialize: }  $\begin{bmatrix}\textcolor{teal}{\vz_1} \\ \textcolor{magenta}{\vz_2} \end{bmatrix} = \mathbf{0}$ \\
            \textbf{Define:}  $h(\vy_1, \vy_2) = \dfrac{\vy_1}{\| \vy_1 \|_2} \cdot \dfrac{\vy_2}{\|\vy_2 \|_2}$
            \State Compute $\begin{bmatrix}\textcolor{teal}{\vz'_1} \\ \textcolor{magenta}{\vz'_2}\end{bmatrix} = \mathrm{FIX}_{f_\vw}\left(\begin{bmatrix}\textcolor{teal}{\vx_1} \\ \textcolor{magenta}{\vx_2} \end{bmatrix}, \begin{bmatrix}\textcolor{teal}{\vz_1} \\ \textcolor{magenta}{\vz_2} \end{bmatrix}\right)$
            \vspace{10pt}
            \\
            \textcolor{gray}{\# Interchange and reinitialize iterates}
            \State Compute $\begin{bmatrix}\textcolor{teal}{\vz''_1} \\ \textcolor{magenta}{\vz''_2} \end{bmatrix} = \mathrm{FIX}_{f_\vw}\left( \begin{bmatrix} \textcolor{teal}{\vx_1} \\ \textcolor{magenta}{\vx_2} \end{bmatrix}, \begin{bmatrix}\textcolor{magenta}{\vz'_2} \\ \textcolor{teal}{\vz'_1} \end{bmatrix}\right)$\\
            \Return $\mathrm{average}(h(\textcolor{teal}{\vz''_1}, \textcolor{teal}{\vz'_1})$, $h(\textcolor{magenta}{\vz''_2}, \textcolor{magenta}{\vz'_2}))$\\
            \end{algorithmic}
        \end{algorithm}

    \end{minipage}
    \hfill
    \begin{minipage}[b]{0.46\textwidth}
        \centering
        \includegraphics[width=0.99\textwidth]{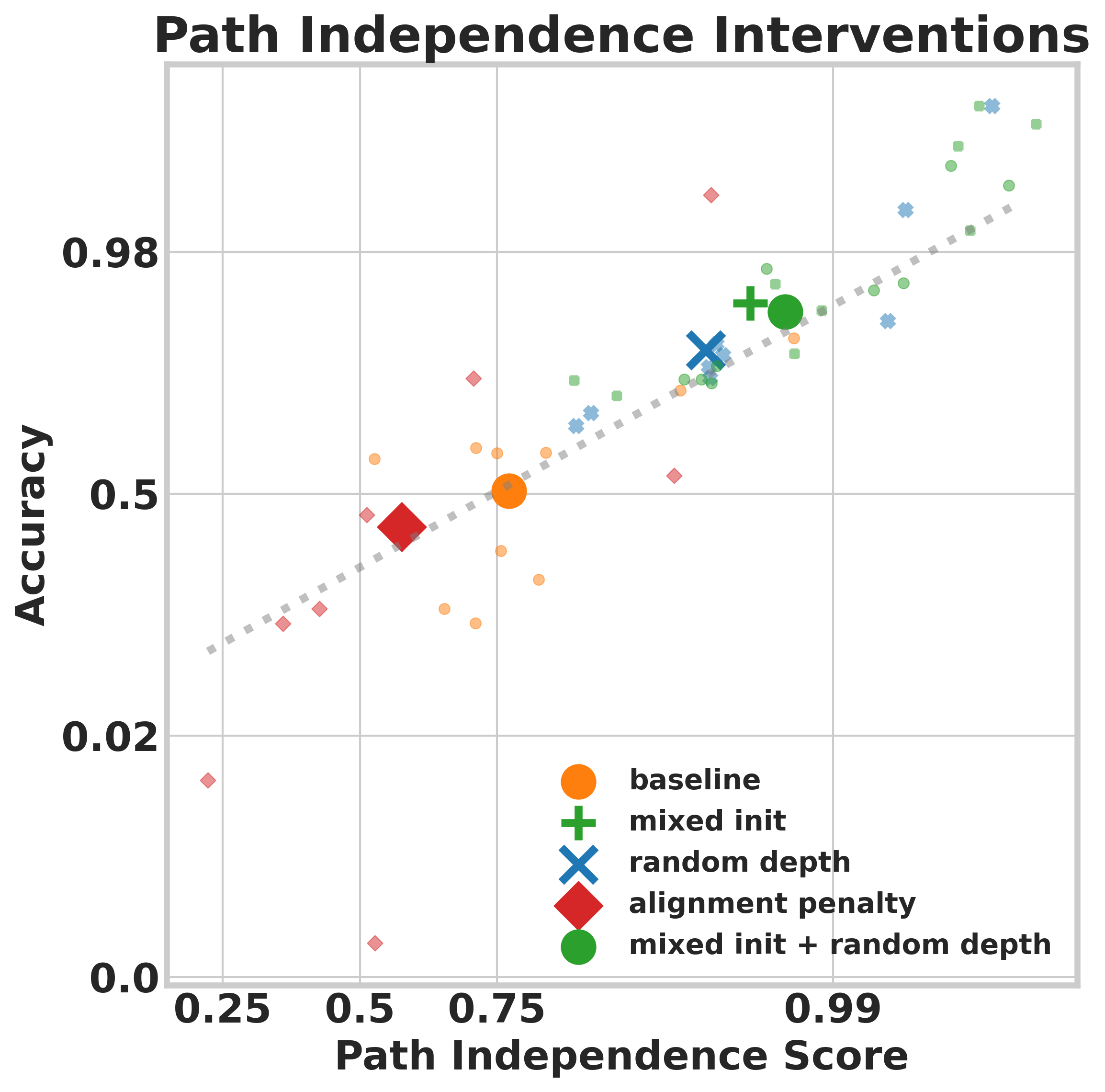}

    \end{minipage}
    \caption{\textbf{(left)} \textbf{AA Score Algorithm: } We provide the algorithm for a simple illustrative case of two inputs. In practice, we consider larger batches. \textbf{(right)} \textbf{Promoting path independence improves generalization in the prefix sum task:} Interventions that are designed to promote path independence (initializing fixed points with random noise or running the fixed point solver with stochastic budget) improves generalization. Conversely, those that hurt path independence (penalty term that directly penalizes fixed point alignment) leads to poorer generalization.}
    \label{mix:algo_and_intervention}
\end{figure}

\label{sec:def_fpa}
We propose a simple metric to quantify path independence based on the directional alignment of the fixed points computed with the same input, but different initializations. We name this metric the \emph{Asymptotic Alignment (AA) score}. Pseudocode to compute the metric is given in \cref{alg:fpa-score}. The AA score is the average cosine similarity between the fixed points obtained with the training time initialization (often simply the zero vector) and the fixed points obtained when one initializes the solver \textit{using the fixed points computed on different inputs}. Higher AA scores (with $1$ being the highest value) imply higher degrees of path independence. In \cref{sec:pi-vs-ood}, we show a strong correlation between path independence and accuracy using the AA score. 

The AA score is cheap to compute, is a reliable indicator of path independence (see below), and is unitless, meaning that networks obtained from different training runs can be compared on equal footing. See the supplementary material for other metrics we've considered for quantifying path independence and why we found AA score to be preferable.

\paragraph{Stress-testing the AA score} 
To stress-test the extent to which the AA score really measures path independence, we search for \textit{adversarial initializations} that are optimized to result in distinct fixed points, hence low AA values. (Unlike adversarial examples, this attack is not constrained to an $\varepsilon$-ball.) We use the L-BFGS \citep{liu1989limited} optimizer, and repeat the search multiple times starting from different fixed point initializations. We include pseudocode in the supplementary material. 

Results of the adversarial stress test can be seen in \cref{table:adversarial-fpa-mazes-prefix-sum}. The results corroborate that the AA score is indeed a reliable measure of path independence; while it isn't possible to find adversarial initializations for high AA score  networks (indicating high path independence), low AA score networks can easily be adversarially initialized to be steered away from the original fixed point estimate.  

\begin{table*}[t!]
\centering
\begin{tabular}{cccccc}
\toprule
Model & Task & AA $\uparrow$ & Accuracy (\%) & Attacked AA $\uparrow$  & Attacked Acc. (\%) \\
\midrule
Non-PI network & Maze & $0.32$ & $87.12$ & $0.09$ & $0$ \\
PI network & Maze & $1.00$ & $100$ & $1.00$ & $100$ \\
\midrule
Non-PI network & Prefix sum & $0.62$ & $66.66$ & $0.18$ & $0$ \\
PI network & Prefix sum & $0.99$ & $100$ & $0.99$ & $100$ \\
\bottomrule
\end{tabular}
\caption{\textbf{Stress-testing the AA Scores:} AA scores for PI vs non-PI networks computed on $13\times13$ mazes and 64 bit prefix sum. Attacked AA refers to the cosine similarity between the fixed point from zero initialization and an adversarial initialization. Non-PI networks can be easily steered away from the initial fixed point estimate through adversarial initializations but it is difficult to do so for PI networks with high AA scores.}
\label{table:adversarial-fpa-mazes-prefix-sum}
\end{table*}

\vspace{-3mm}
\section{Path Independence Correlates with Upward Generalization} \label{sec:pi-vs-ood}
\vspace{-1mm}
\label{sec:pi_corr_ood}
Is path independence (as measured by the AA score) a strong predictor of upwards generalization? We took the trained networks from \cref{sec:algo_ood_eq}, computed their average AA scores on in- and out-of-distribution splits and inspected whether the AA scores are correlated with upward generalization.

On prefix sum experiments, we varied 1) network depth, 2) whether or not weight norm (wnorm) \citep{DBLP:journals/corr/SalimansK16} was used or not,\footnote{\citet{bai2019deep} report that weight norm helps stabilize the training of DEQ models.} 3) learning rate (one of [0.01, 0.001, 0.0001]), 4) forward solver (fixed point iterations or Anderson acceleration \citep{anderson1965iterative}, and 5) the gradient estimator (backprop or implicit gradients).\footnote{Note that the deep equilibrium model (DEQ) setup~\citep{bai2019deep} correspond to using a root solver (such as Anderson) for the forward pass and implicit gradients for the backward pass.} On the maze experiments, we varied 1) network depth, 2) use of  weight norm, 3) forward solver (fixed point iterations or Broyden solver \citep{broyden1965class}), and 5) the gradient estimator (backprop or implicit gradients).

\begin{figure*}[!]
\centering
\begin{subfigure}[t]{0.49\textwidth}
  \includegraphics[width=1\textwidth]{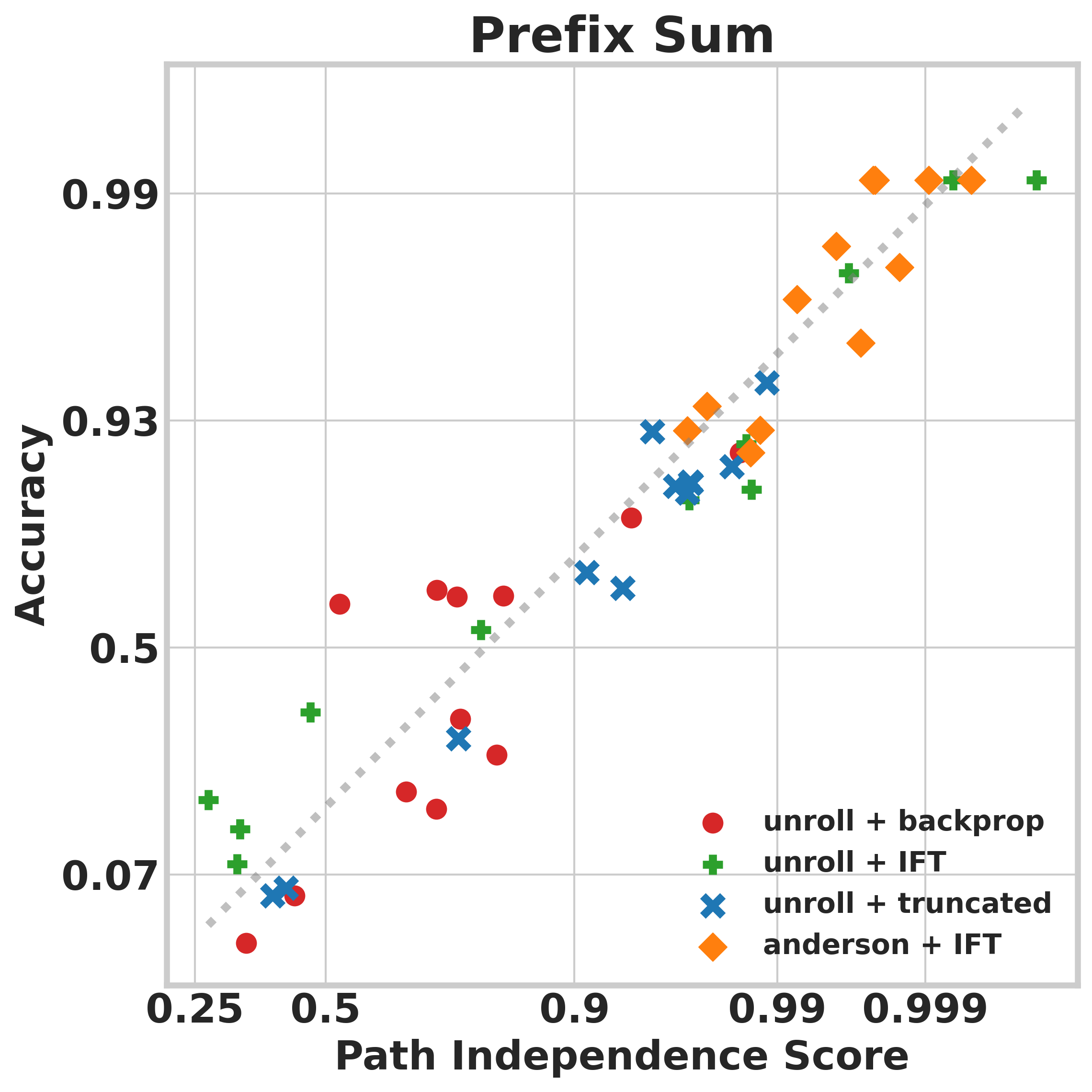}
  \caption{Prefix Sum}
  \label{fig:pi-corr-prefix-sum}
\end{subfigure}
\begin{subfigure}[t]{0.49\textwidth}
  \includegraphics[width=1\textwidth]{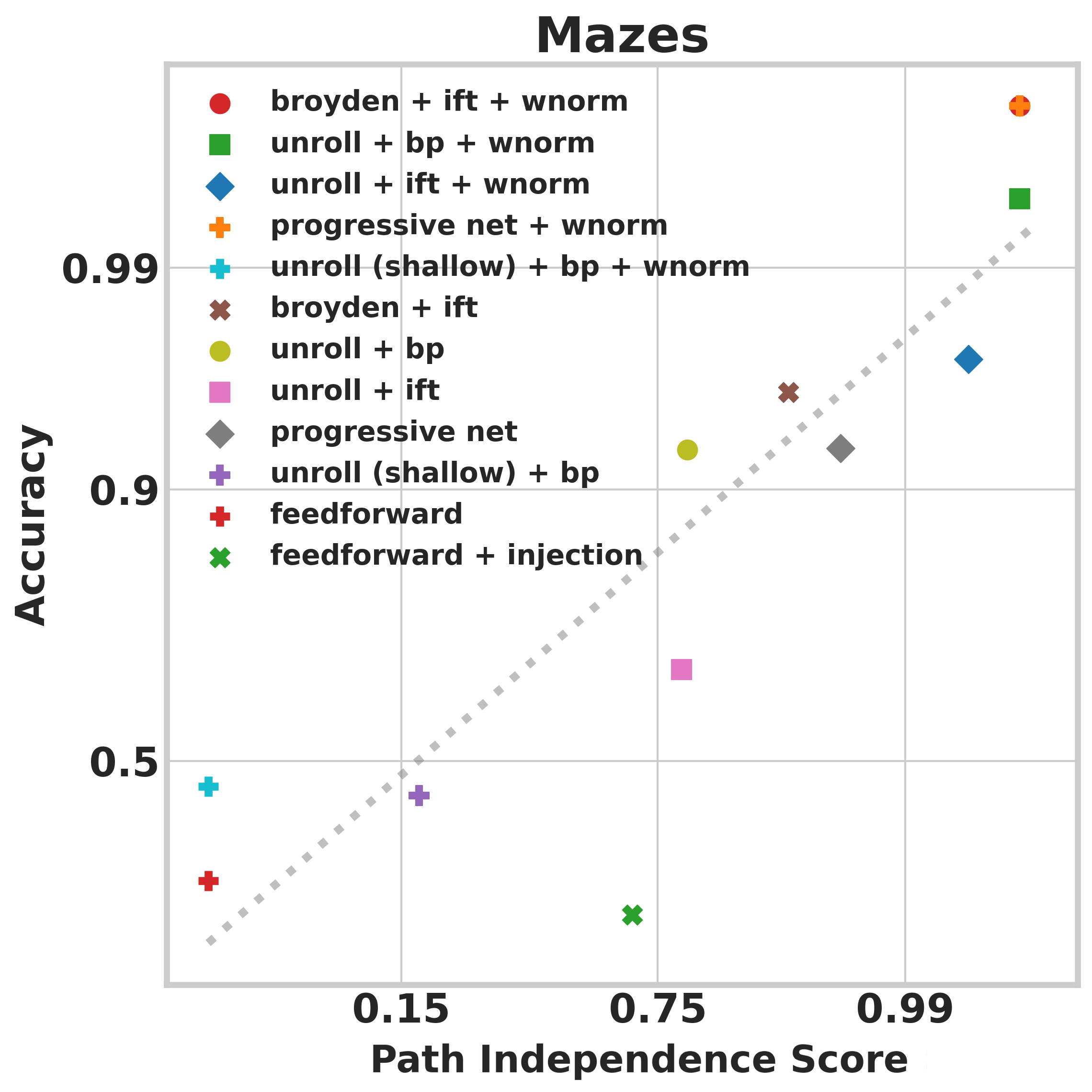}
  \caption{Mazes}
  \label{fig:pi-corr-mazes}
\end{subfigure}
\caption{High AA scores correlate with good upward generalization. For a given choice of an architecture and a task, the reported numbers are averaged over problem instances of different dimensions. We apply the probit transformation along both axes, following \citet{miller2021accuracy}. Accuracies and AA scores are capped at 0.999 for compatibility with the probit transform. } %
\label{fig:pi_corr_acc}
\end{figure*}

\cref{fig:pi_corr_acc} displays our findings. We evaluated performance on a mixture of in- and OOD validation data; results on individual data splits can be found in the supplementary material. The results show a strong correlation between AA score and accuracy when the inference depth is large enough. This shows that PI networks allow for scaling test-time compute to improve test-time accuracy (see also \cref{fig:pi-vs-nonpi-accuracy}). The in-distribution validation performance of non-PI networks degrades with deeper inference depths. Unsurprisingly, these networks generalize poorly on harder problem instances that require deeper inference depths (i.e. problem instances that provably require at least a given number of layers to handle). Further results on the BlurryMNIST, matrix inversion and edge copy tasks can be found in Supplementary Material \ref{app:bmnist}, \ref{app:minvert} and \ref{app:ecopy}. 
\looseness=-1

\section{Experimental Manipulations of Path Independence}

\label{sec:manipulations}

The previous section demonstrates a strong correlation between path independence and the ability to exploit additional test-time iterations. Unfortunately, we can't make a causal claim based on these studies: the observed effect could have been due to an unobserved confounder. 
In this section, we intervene directly on path independence by imposing regularizers which directly encourage or penalize path independence. 
We find that interventions designed to \textit{promote} path independence also improve generalization, while interventions designed to \textit{reduce} path independence also hurt generalization. 
\looseness=-1

\subsection{Promoting Path Independence via Randomized Forward Passes}
\label{sec:positive_interv}

A straightforward way to encourage path independence is simply to initialize the hidden states with random noise during training. To this end, we experimented with initializing the hidden states with zeros on half of the examples in the batch, and with standard Gaussian noise on the rest of the examples. The reason to include the zero-initializations at training time is that we initialize from zeros at test time - not including this initialization during traning time causes a distribution shift.

Another way to promote path independence is simply running the forward solver with randomized compute budgets/depths during training time. While a path independent solution can be expected to be robust against this intervention, a path dependent one will fail. 

We took the training configurations of the 12 prefix-sum networks described in \cref{sec:pi_corr_ood} that use fixed point iterations in their forward pass, and backpropagation gradient in their backward pass, and retrained them separately with the aforementioned mixed initialization and randomized depth strategies without modifying any other experimental conditions. As can be seen in \cref{mix:algo_and_intervention}, the interventions lead to strong test-time path independent neural networks, while also reliably improving in- and out-of-distribution validation accuracy. We especially emphasize that shallow networks trained with mixed initialization actually remain far from having high AA scores using the training-time forward pass conditions due to lack of convergence. However, since the mixed initialization strategy results in path independent networks, scaling up test-time compute budget leads to high AA scores, and therefore high upwards generalization. 

\subsection{Penalizing Path Independence via the Fixed Point Alignment Penalty}
Does an intervention that results in less path independence also result in poorer upwards generalization? Like in the mixed initialization experiment, we retrained the 12 unroll + backpropagation networks with an additional auxiliary loss term that penalizes the dot product between the fixed points computed from the same input, but different initializations sampled from standard Gaussian noise. \cref{mix:algo_and_intervention} shows that this intervention succeeded in pushing the AA scores down, while also keeping the accuracy on the same trend line. 

\section{Disambiguating Convergence and Path Independence} \label{sec:disambiguate-convergence-pi}

Is convergence necessary for path independence?
We answer this statement in the negative, and show that neither training-time convergence nor test-time convergence is required for path independence. 
Instead convergence to the same limiting behavior regardless of initialization is important.

\paragraph{Training Time Convergence}
We consider two implicitly trained equilibrium models trained on the mazes task---one trained with implicit gradients computed via implicit function theorem (IFT), and the other trained with an approximation of the (inverse) Jacobian, called  phantom gradients \citep{geng2021training}. We report the values of residuals (\ie $\| f_\vw (\vx, \vz) - \vz \|_2$), AA scores and accuracies observed for in- and out-of-distribution data for mazes in \cref{table:training-time-convergence}. We observe that DEQs trained with phantom gradients have higher values of in-distribution residuals but are path independent, as indicated by their high AA scores, and show strong upward generalization as indicated by their good accuracy. 

The mixed-initialization intervention described in \cref{sec:positive_interv} also leads to a separation between training-time convergence and path independence. We found that it is possible to train very shallow (\ie 5 layer) unrolled networks that, while being very far from converging during training and attaining poor in-distribution generalization, are able to converge and achieve perfect performance when run for many more iterations during test time. Details are provided in the supplementary material.

\begin{table*}[t!]
\centering
\begin{tabular}{ccccccc}
\toprule
\multirow{2}{*}{Model} & \multicolumn{2}{c}{Residual $\downarrow$} & \multicolumn{2}{c}{AA score $\uparrow$} & \multicolumn{2}{c}{Accuracy (\%) $\uparrow$} \\
\cmidrule{2-3} \cmidrule{4-5} \cmidrule{6-7}
& In-dist & OOD & In-dist & OOD & In-dist & OOD \\
\midrule
DEQ (phantom grad.) & $11.83$ & $0.016$ & $0.96$ & $0.99$ & $99.96$ & $99.88$\\ %
DEQ (IFT)& $1.4$ & $0.011$ & $0.99$ & $0.99$ & $99.99$ & $100$ \\  %
\bottomrule
\end{tabular}
\caption{Training-time convergence is not needed for path independence: models might show poor training-time convergence (as shown by high values of residuals) but still be path independent. Residual, AA score, and Accuracy for DEQ trained with IFT vs phantom  gradients. In-distribution (In-dist) results were computed on $9 \times 9$ mazes, and OOD results were computed on $25\times25$ mazes.}
\label{table:training-time-convergence}
\vspace{4mm}
\end{table*}

\textbf{Test Time Convergence} From \cref{table:training-time-convergence}, one might conclude that test time convergence is important for path independence. However, we show that this %
connection is not necessary, and convergence to the same fixed point is not a required condition for path independence. We study test time convergence properties of an unrolled weight-tied input-injected network trained with backpropagation under different solvers. This network is highly path independent using either the Broyden solver or fixed point iterations, as indicated by its high AA scores (0.99) on both in- and out-of-distribution data. We visualize the values of test-time residuals with fixed point iterations and Broyden's method in \cref{fig:test-time-residuals}. Both these solvers converge to different limit cycles but still show good upward generalization.

\section{Path Independence on a Per-Example Level}
\label{sec:per_instance_analysis}
\begin{figure}[!t]
    \centering
    \begin{minipage}[b]{0.49\textwidth}
        \centering
        \includegraphics[width=0.8\textwidth]{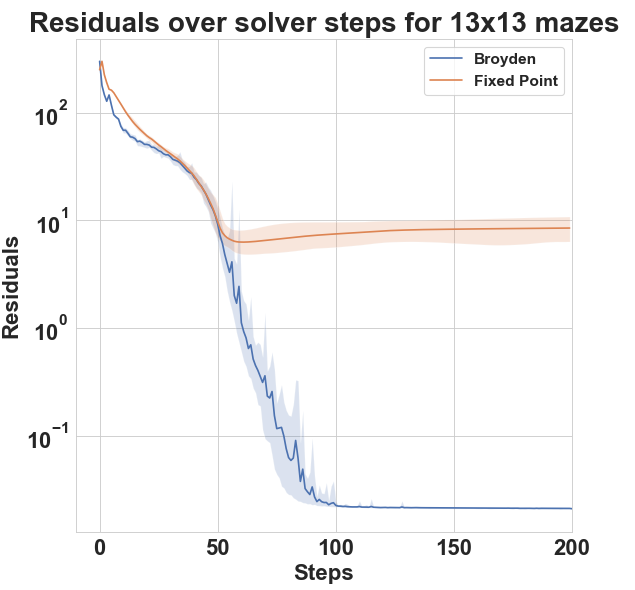}
    \subcaption[]{\label{fig:test-time-residuals}}
    \end{minipage}
    \hfill
    \begin{minipage}[b]{0.49\textwidth}
    \centering
    \includegraphics[width=0.9\textwidth]{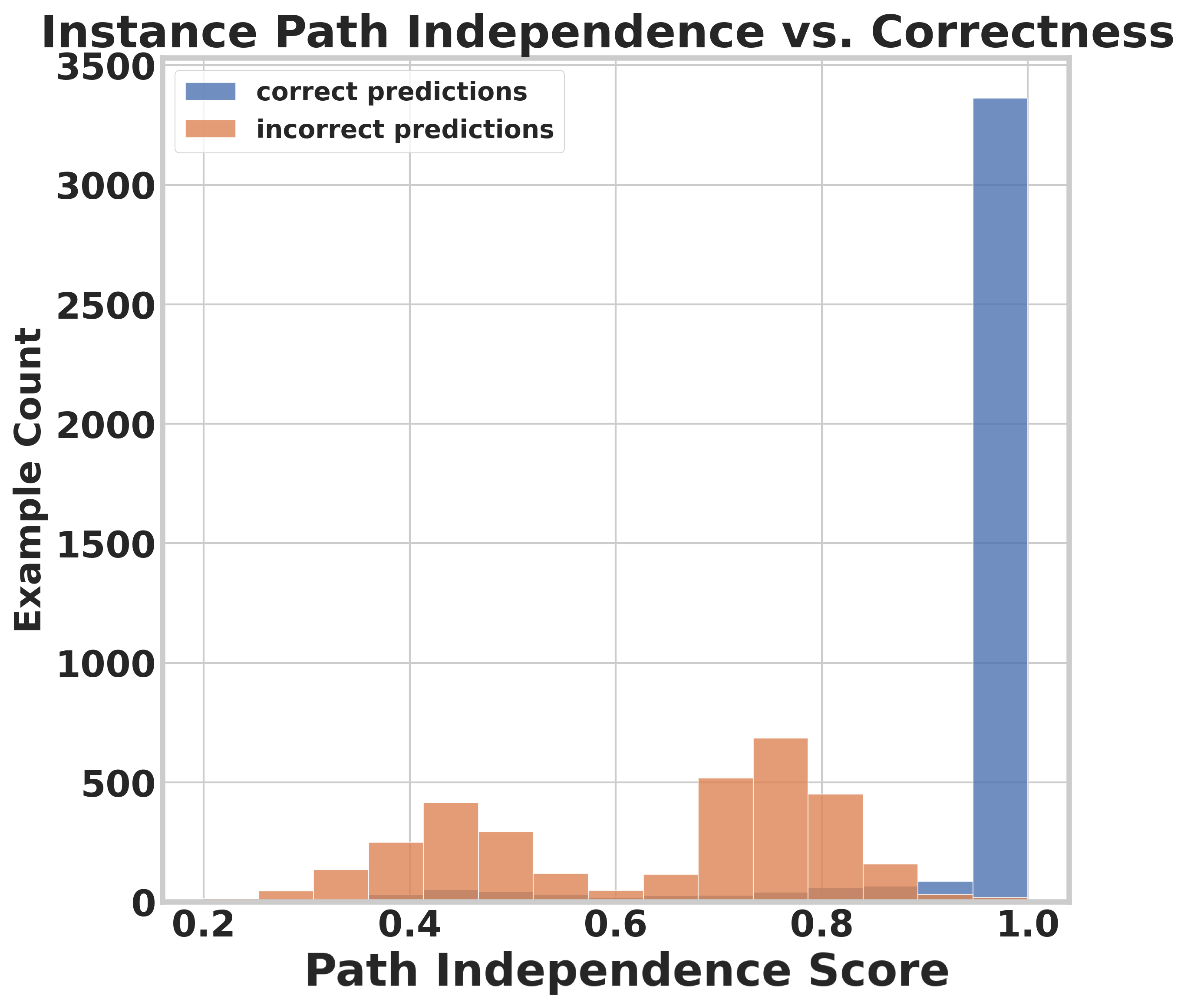}
    \subcaption[]{\label{fig:pis-corr-incorr-ver}}
    \end{minipage}
    \caption{(\textbf{Left}) Different solvers display differing asymptotic behaviours but still achieve good upwards  generalization. Here, the network has an adversarial AA score of 0.99, and achieves accuracy of 99.98\% (fixed point iterations) and 99.97\% (Broyden solver) respectively on the mazes task; (\textbf{Right}) Per-instance path independence is highly correlated with correctness of predictions for prefix sum task. }
\end{figure}

The connection between path independence and prediction correctness also largely holds on a per-instance basis. Using the prefix-sum networks trained with the mixed-initialization strategy (the most performant group of networks in our intervention experiments), we plotted the distribution of per-instance fixed point alignment scores, colored by whether the prediction on that instance was correct or not in Figure  \ref{fig:pis-corr-incorr-ver}. This suggests that path independence can be used as a valuable sanity-check to determine whether a prediction is correct or not without the need for any label data, both in- and out-of-distribution. We provide a more in-depth per-instance analysis in the supplementary material.
\looseness=-1

\section{Related Work}
\label{sec:related}

There is a long line of research on neural networks that can adapt their computational budget based on the complexity of the task they are learning to solve---akin to the intrinsic mechanism in humans to reason and solve problems. \citet{schmidhuber2012self} introduced self-delimiting neural networks which are a type of recurrent neural networks (RNNs) that adapt their compute based on the output of a special "halt" neuron. Adaptive computation time (ACT) \citep{DBLP:journals/corr/Graves16} also uses the output of a sigmoidal halting unit to determine the termination condition of an RNN, but it avoids long "thinking" time by explicitly penalizing it. Subsequent works have successfully applied variants of ACT in image classification and object detection \citep{figurnov2017spatially}, visual reasoning \citep{eyzaguirre2020differentiable}, Transformers \citep{vaswani2017attention} for language modelling \citep{dehghani2019,Elbayad2020Depth-Adaptive,liu2021faster}, and recognizing textual entailment \citep{neumann2016learning}. PonderNet \citep{banino2021pondernet} reformulates the halting policy of ACT as a probabilistic model, and adds a regularization term in the loss objective to encourage exploration. With these additions, PonderNet can extrapolate to more difficult examples on the parity task, first proposed by \citet{graves2016adaptive}: in a vector with entries of 0, -1, and 1, output 1 for odd number of ones, and 0 otherwise. In this work, we do not optimize or penalize the network for the number of computational steps. Our main goal is to understand the underlying mechanism that results in scalable  generalization of equilibrium models on harder problem instances. Our current work is closely related to previous work by \citet{schwarzschild2021can} and \citep{bansal2022end} that propose architectural choices and training mechanisms that enable weight tied networks to generalize on harder problem instances.  We relate these papers' contributions to ours in Section \ref{sec:pi_necessary}. 

Another family of models with the property of adaptive inference compute budget is early exit networks \citep{teerapittayanon2016branchynet, laskaridis2021adaptive}. These networks have multiple additional “exit” prediction heads along their depth. At inference time, the result that satisfies an exit policy is selected as the prediction output. %
This approach of designing adaptive networks has been  adapted both in natural language processing \citep{schwartz2020right, DBLP:journals/corr/abs-2005-02534, Elbayad2020DepthAdaptiveT, zhou2020bert, liu2020fastbert} and vision \citep{li2017not, wang2018skipnet, xing2020early, kouris2021multi}. Most of these architectures have  complex sub-modules that are trained in multiple stages, and require complex exit policies. In contrast, equilibrium models have a simple architecture, and can use root solvers to efficiently solve for the fixed point at inference.

More complex transformer-based language models like GPT-3 also struggle to generalize well on simple algorithmic tasks like addition \citep{brown2020language}.
Recent work by \citet{nye2021show} shows that transformers can be trained to perform well on algorithmic tasks and generalize on OOD data by emitting the intermediate steps of an algorithm to a buffer called ``scratchpad''. Using a scratchpad enables the model to revisit its errors and correct them.

Path independence is closely related to the concept of \textit{global stability and global convergence} in control theory and optimization. This concept is somewhat overloaded, as it sometimes requires convergence to a single point~\citep{slotine1991applied}, and sometimes implies the system is convergent everywhere, even if to different points \citep{wang2003global, sriperumbudur2009convergence}. We thus choose the term \emph{path independence} to refer specifically to the fact that the system will converge to the same limiting behavior (whatever that might be) regardless of the initial state of the system.

\section{Conclusion}
Being able to attain better levels of performance using a larger inference-time compute budget is a feat that eludes most standard deep learning architectures. This is especially relevant for tasks that require \textit{upwards generalization}, \ie the ability to generalize from easy problem instances to hard ones. We show that equilibrium models are capable of displaying upwards generalization by exploiting scalable test-time compute. We link this to a phenomenon we call \textit{path independence}: the tendency of an equilibrium network to converge to the same limiting behavior given an input, regardless of the initial conditions. We investigate this phenomenon through careful experiments and verify that  path independent networks indeed generalize well on harder problem instances by exploiting more test time compute. Moreover, interventions on training conditions that promote path independence also improve upwards generalization, while those that penalize it hurt this capability. Our findings suggest that path independent equilibrium models are a promising direction towards building general purpose learning systems whose test-time performance improves with more compute. 
\newpage

\section{Acknowledgements}
AP is supported by a grant from the Bosch Center for Artificial Intelligence. JT acknowledges support from the Center on Long-Term Risk Fund.
CA is supported by NSERC Canada Graduate School - Doctorate scholarship. 
\bibliographystyle{plainnat}
\bibliography{references}

\begin{thebibliography}{57}
\providecommand{\natexlab}[1]{#1}
\providecommand{\url}[1]{\texttt{#1}}
\expandafter\ifx\csname urlstyle\endcsname\relax
  \providecommand{\doi}[1]{doi: #1}\else
  \providecommand{\doi}{doi: \begingroup \urlstyle{rm}\Url}\fi

\bibitem[Anderson(1965)]{anderson1965iterative}
Donald~G Anderson.
\newblock Iterative procedures for nonlinear integral equations.
\newblock \emph{Journal of the ACM (JACM)}, 12\penalty0 (4):\penalty0 547--560,
  1965.

\bibitem[Ba et~al.(2016)Ba, Kiros, and Hinton]{ba2016layer}
Jimmy~Lei Ba, Jamie~Ryan Kiros, and Geoffrey~E Hinton.
\newblock Layer normalization.
\newblock \emph{arXiv preprint arXiv:1607.06450}, 2016.

\bibitem[Bai et~al.(2019)Bai, Kolter, and Koltun]{bai2019deep}
Shaojie Bai, J.~Zico Kolter, and Vladlen Koltun.
\newblock Deep equilibrium models.
\newblock In \emph{Neural Information Processing Systems}, 2019.

\bibitem[Bai et~al.(2020)Bai, Koltun, and Kolter]{bai2020multiscale}
Shaojie Bai, Vladlen Koltun, and J.~Zico Kolter.
\newblock Multiscale deep equilibrium models.
\newblock In \emph{Neural Information Processing Systems}, 2020.

\bibitem[Bai et~al.(2021)Bai, Koltun, and Kolter]{Bai2021StabilizingEM}
Shaojie Bai, Vladlen Koltun, and J.~Zico Kolter.
\newblock Stabilizing equilibrium models by jacobian regularization.
\newblock In \emph{ICML}, 2021.

\bibitem[Banino et~al.(2021)Banino, Balaguer, and
  Blundell]{banino2021pondernet}
Andrea Banino, Jan Balaguer, and Charles Blundell.
\newblock Pondernet: Learning to ponder.
\newblock \emph{arXiv preprint arXiv:2107.05407}, 2021.

\bibitem[Bansal et~al.(2022)Bansal, Schwarzschild, Borgnia, Emam, Huang,
  Goldblum, and Goldstein]{bansal2022end}
Arpit Bansal, Avi Schwarzschild, Eitan Borgnia, Zeyad Emam, Furong Huang, Micah
  Goldblum, and Tom Goldstein.
\newblock End-to-end algorithm synthesis with recurrent networks: Logical
  extrapolation without overthinking.
\newblock \emph{arXiv preprint arXiv:2202.05826}, 2022.

\bibitem[Brown et~al.(2020)Brown, Mann, Ryder, Subbiah, Kaplan, Dhariwal,
  Neelakantan, Shyam, Sastry, Askell, et~al.]{brown2020language}
Tom Brown, Benjamin Mann, Nick Ryder, Melanie Subbiah, Jared~D Kaplan, Prafulla
  Dhariwal, Arvind Neelakantan, Pranav Shyam, Girish Sastry, Amanda Askell,
  et~al.
\newblock Language models are few-shot learners.
\newblock \emph{Advances in neural information processing systems},
  33:\penalty0 1877--1901, 2020.

\bibitem[Broyden(1965)]{broyden1965class}
Charles~G Broyden.
\newblock A class of methods for solving nonlinear simultaneous equations.
\newblock \emph{Mathematics of Computation}, 1965.

\bibitem[Dehghani et~al.(2019)Dehghani, Gouws, Vinyals, Uszkoreit, and
  Kaiser]{dehghani2019}
Mostafa Dehghani, Stephan Gouws, Oriol Vinyals, Jakob Uszkoreit, and Lukasz
  Kaiser.
\newblock Universal transformers.
\newblock 2019.
\newblock URL \url{https://openreview.net/pdf?id=HyzdRiR9Y7}.

\bibitem[Du et~al.(2022)Du, Li, Tenenbaum, and Mordatch]{du2022learning}
Yilun Du, Shuang Li, Joshua Tenenbaum, and Igor Mordatch.
\newblock Learning iterative reasoning through energy minimization.
\newblock In \emph{International Conference on Machine Learning}, pages
  5570--5582. PMLR, 2022.

\bibitem[Elbayad et~al.(2020{\natexlab{a}})Elbayad, Gu, Grave, and
  Auli]{Elbayad2020Depth-Adaptive}
Maha Elbayad, Jiatao Gu, Edouard Grave, and Michael Auli.
\newblock Depth-adaptive transformer.
\newblock In \emph{International Conference on Learning Representations},
  2020{\natexlab{a}}.
\newblock URL \url{https://openreview.net/forum?id=SJg7KhVKPH}.

\bibitem[Elbayad et~al.(2020{\natexlab{b}})Elbayad, Gu, Grave, and
  Auli]{Elbayad2020DepthAdaptiveT}
Maha Elbayad, Jiatao Gu, Edouard Grave, and Michael Auli.
\newblock Depth-adaptive transformer.
\newblock \emph{ArXiv}, abs/1910.10073, 2020{\natexlab{b}}.

\bibitem[Eyzaguirre and Soto(2020)]{eyzaguirre2020differentiable}
Cristobal Eyzaguirre and Alvaro Soto.
\newblock Differentiable adaptive computation time for visual reasoning.
\newblock In \emph{Proceedings of the ieee/cvf conference on computer vision
  and pattern recognition}, pages 12817--12825, 2020.

\bibitem[Figurnov et~al.(2017)Figurnov, Collins, Zhu, Zhang, Huang, Vetrov, and
  Salakhutdinov]{figurnov2017spatially}
Michael Figurnov, Maxwell~D Collins, Yukun Zhu, Li~Zhang, Jonathan Huang,
  Dmitry Vetrov, and Ruslan Salakhutdinov.
\newblock Spatially adaptive computation time for residual networks.
\newblock In \emph{Proceedings of the IEEE Conference on Computer Vision and
  Pattern Recognition}, pages 1039--1048, 2017.

\bibitem[Fung et~al.(2021)Fung, Heaton, Li, McKenzie, Osher, and
  Yin]{fung2021fixed}
Samy~Wu Fung, Howard Heaton, Qiuwei Li, Daniel McKenzie, Stanley Osher, and
  Wotao Yin.
\newblock Fixed point networks: Implicit depth models with jacobian-free
  backprop.
\newblock \emph{arXiv e-prints}, pages arXiv--2103, 2021.

\bibitem[Geng et~al.(2021{\natexlab{a}})Geng, Guo, Chen, Li, Wei, and
  Lin]{geng2021attention}
Zhengyang Geng, Meng-Hao Guo, Hongxu Chen, Xia Li, Ke~Wei, and Zhouchen Lin.
\newblock Is attention better than matrix decomposition?
\newblock In \emph{International Conference on Learning Representations
  (ICLR)}, 2021{\natexlab{a}}.

\bibitem[Geng et~al.(2021{\natexlab{b}})Geng, Zhang, Bai, Wang, and
  Lin]{geng2021training}
Zhengyang Geng, Xin-Yu Zhang, Shaojie Bai, Yisen Wang, and Zhouchen Lin.
\newblock On training implicit models.
\newblock \emph{Advances in Neural Information Processing Systems}, 34,
  2021{\natexlab{b}}.

\bibitem[Graves(2016{\natexlab{a}})]{DBLP:journals/corr/Graves16}
Alex Graves.
\newblock Adaptive computation time for recurrent neural networks.
\newblock \emph{CoRR}, abs/1603.08983, 2016{\natexlab{a}}.
\newblock URL \url{http://arxiv.org/abs/1603.08983}.

\bibitem[Graves(2016{\natexlab{b}})]{graves2016adaptive}
Alex Graves.
\newblock Adaptive computation time for recurrent neural networks.
\newblock \emph{arXiv preprint arXiv:1603.08983}, 2016{\natexlab{b}}.

\bibitem[Gu et~al.(2020)Gu, Chang, Zhu, Sojoudi, and El~Ghaoui]{gu2020implicit}
Fangda Gu, Heng Chang, Wenwu Zhu, Somayeh Sojoudi, and Laurent El~Ghaoui.
\newblock Implicit graph neural networks.
\newblock \emph{Advances in Neural Information Processing Systems},
  33:\penalty0 11984--11995, 2020.

\bibitem[Hu et~al.(2019)Hu, Liu, Gomes, Zitnik, Liang, Pande, and
  Leskovec]{hu2019strategies}
Weihua Hu, Bowen Liu, Joseph Gomes, Marinka Zitnik, Percy Liang, Vijay Pande,
  and Jure Leskovec.
\newblock Strategies for pre-training graph neural networks.
\newblock \emph{arXiv preprint arXiv:1905.12265}, 2019.

\bibitem[Kingma and Ba(2014)]{kingma2014adam}
Diederik~P Kingma and Jimmy Ba.
\newblock Adam: A method for stochastic optimization.
\newblock \emph{arXiv preprint arXiv:1412.6980}, 2014.

\bibitem[Koh et~al.(2021)Koh, Sagawa, Marklund, Xie, Zhang, Balsubramani, Hu,
  Yasunaga, Phillips, Gao, et~al.]{koh2021wilds}
Pang~Wei Koh, Shiori Sagawa, Henrik Marklund, Sang~Michael Xie, Marvin Zhang,
  Akshay Balsubramani, Weihua Hu, Michihiro Yasunaga, Richard~Lanas Phillips,
  Irena Gao, et~al.
\newblock Wilds: A benchmark of in-the-wild distribution shifts.
\newblock In \emph{International Conference on Machine Learning}, pages
  5637--5664. PMLR, 2021.

\bibitem[Kolter et~al.(2020)Kolter, Duvenaud, and Johnson]{kolter2020tutorial}
Zico Kolter, David Duvenaud, and Matthew Johnson.
\newblock Tutorial: Deep implicit layers-neural odes, deep equilibirum models,
  and beyond, 2020.

\bibitem[Kouris et~al.(2021)Kouris, Venieris, Laskaridis, and
  Lane]{kouris2021multi}
Alexandros Kouris, Stylianos~I Venieris, Stefanos Laskaridis, and Nicholas~D
  Lane.
\newblock Multi-exit semantic segmentation networks.
\newblock \emph{arXiv preprint arXiv:2106.03527}, 2021.

\bibitem[Laskaridis et~al.(2021)Laskaridis, Kouris, and
  Lane]{laskaridis2021adaptive}
Stefanos Laskaridis, Alexandros Kouris, and Nicholas~D Lane.
\newblock Adaptive inference through early-exit networks: Design, challenges
  and directions.
\newblock In \emph{Proceedings of the 5th International Workshop on Embedded
  and Mobile Deep Learning}, pages 1--6, 2021.

\bibitem[Li et~al.(2017)Li, Liu, Luo, Change~Loy, and Tang]{li2017not}
Xiaoxiao Li, Ziwei Liu, Ping Luo, Chen Change~Loy, and Xiaoou Tang.
\newblock Not all pixels are equal: Difficulty-aware semantic segmentation via
  deep layer cascade.
\newblock In \emph{Proceedings of the IEEE conference on computer vision and
  pattern recognition}, pages 3193--3202, 2017.

\bibitem[Liang et~al.()Liang, Anil, Wu, and Grosse]{liangout}
Kaiqu Liang, Cem Anil, Yuhuai Wu, and Roger Grosse.
\newblock Out-of-distribution generalization with deep equilibrium models.
\newblock \emph{Uncertainty and Robustness in Deep Learning, ICML 2021}.

\bibitem[Liao et~al.(2018)Liao, Xiong, Fetaya, Zhang, Yoon, Pitkow, Urtasun,
  and Zemel]{liao2018reviving}
Renjie Liao, Yuwen Xiong, Ethan Fetaya, Lisa Zhang, KiJung Yoon, Xaq Pitkow,
  Raquel Urtasun, and Richard Zemel.
\newblock Reviving and improving recurrent back-propagation.
\newblock In \emph{International Conference on Machine Learning}, pages
  3082--3091. PMLR, 2018.

\bibitem[Liu and Nocedal(1989)]{liu1989limited}
Dong~C Liu and Jorge Nocedal.
\newblock On the limited memory bfgs method for large scale optimization.
\newblock \emph{Mathematical programming}, 45\penalty0 (1):\penalty0 503--528,
  1989.

\bibitem[Liu et~al.(2021{\natexlab{a}})Liu, Kawaguchi, Hooi, Wang, and
  Xiao]{Liu2021EIGNNEI}
Juncheng Liu, Kenji Kawaguchi, Bryan Hooi, Yiwei Wang, and X.~Xiao.
\newblock Eignn: Efficient infinite-depth graph neural networks.
\newblock In \emph{NeurIPS}, 2021{\natexlab{a}}.

\bibitem[Liu et~al.(2020)Liu, Zhou, Zhao, Wang, Deng, and Ju]{liu2020fastbert}
Weijie Liu, Peng Zhou, Zhe Zhao, Zhiruo Wang, Haotang Deng, and Qi~Ju.
\newblock Fastbert: a self-distilling bert with adaptive inference time.
\newblock \emph{arXiv preprint arXiv:2004.02178}, 2020.

\bibitem[Liu et~al.(2021{\natexlab{b}})Liu, Meng, Zhou, Chen, and
  Xu]{liu2021faster}
Yijin Liu, Fandong Meng, Jie Zhou, Yufeng Chen, and Jinan Xu.
\newblock Faster depth-adaptive transformers.
\newblock In \emph{Proceedings of the AAAI Conference on Artificial
  Intelligence}, volume~35, pages 13424--13432, 2021{\natexlab{b}}.

\bibitem[McClelland and Rumelhart(1989)]{mcclelland1989explorations}
James~L McClelland and David~E Rumelhart.
\newblock \emph{Explorations in parallel distributed processing: A handbook of
  models, programs, and exercises}.
\newblock MIT press, 1989.

\bibitem[Miller et~al.(2021)Miller, Taori, Raghunathan, Sagawa, Koh, Shankar,
  Liang, Carmon, and Schmidt]{miller2021accuracy}
John~P Miller, Rohan Taori, Aditi Raghunathan, Shiori Sagawa, Pang~Wei Koh,
  Vaishaal Shankar, Percy Liang, Yair Carmon, and Ludwig Schmidt.
\newblock Accuracy on the line: on the strong correlation between
  out-of-distribution and in-distribution generalization.
\newblock In \emph{International Conference on Machine Learning}, pages
  7721--7735. PMLR, 2021.

\bibitem[Neumann et~al.(2016)Neumann, Stenetorp, and
  Riedel]{neumann2016learning}
Mark Neumann, Pontus Stenetorp, and Sebastian Riedel.
\newblock Learning to reason with adaptive computation.
\newblock \emph{arXiv preprint arXiv:1610.07647}, 2016.

\bibitem[Nye et~al.(2021)Nye, Andreassen, Gur-Ari, Michalewski, Austin, Bieber,
  Dohan, Lewkowycz, Bosma, Luan, et~al.]{nye2021show}
Maxwell Nye, Anders~Johan Andreassen, Guy Gur-Ari, Henryk Michalewski, Jacob
  Austin, David Bieber, David Dohan, Aitor Lewkowycz, Maarten Bosma, David
  Luan, et~al.
\newblock Show your work: Scratchpads for intermediate computation with
  language models.
\newblock \emph{arXiv preprint arXiv:2112.00114}, 2021.

\bibitem[Revay et~al.(2020)Revay, Wang, and Manchester]{revay2020lipschitz}
Max Revay, Ruigang Wang, and Ian~R Manchester.
\newblock Lipschitz bounded equilibrium networks.
\newblock \emph{arXiv:2010.01732}, 2020.

\bibitem[Salimans and Kingma(2016)]{DBLP:journals/corr/SalimansK16}
Tim Salimans and Diederik~P. Kingma.
\newblock Weight normalization: {A} simple reparameterization to accelerate
  training of deep neural networks.
\newblock \emph{CoRR}, abs/1602.07868, 2016.
\newblock URL \url{http://arxiv.org/abs/1602.07868}.

\bibitem[Schmidhuber(2012)]{schmidhuber2012self}
J{\"u}rgen Schmidhuber.
\newblock Self-delimiting neural networks.
\newblock \emph{arXiv preprint arXiv:1210.0118}, 2012.

\bibitem[Schwartz et~al.(2020)Schwartz, Stanovsky, Swayamdipta, Dodge, and
  Smith]{schwartz2020right}
Roy Schwartz, Gabriel Stanovsky, Swabha Swayamdipta, Jesse Dodge, and Noah~A
  Smith.
\newblock The right tool for the job: Matching model and instance complexities.
\newblock \emph{arXiv preprint arXiv:2004.07453}, 2020.

\bibitem[Schwarzschild et~al.(2021{\natexlab{a}})Schwarzschild, Borgnia, Gupta,
  Bansal, Emam, Huang, Goldblum, and Goldstein]{schwarzschild2021datasets}
Avi Schwarzschild, Eitan Borgnia, Arjun Gupta, Arpit Bansal, Zeyad Emam, Furong
  Huang, Micah Goldblum, and Tom Goldstein.
\newblock Datasets for studying generalization from easy to hard examples.
\newblock \emph{arXiv preprint arXiv:2108.06011}, 2021{\natexlab{a}}.

\bibitem[Schwarzschild et~al.(2021{\natexlab{b}})Schwarzschild, Borgnia, Gupta,
  Huang, Vishkin, Goldblum, and Goldstein]{schwarzschild2021can}
Avi Schwarzschild, Eitan Borgnia, Arjun Gupta, Furong Huang, Uzi Vishkin, Micah
  Goldblum, and Tom Goldstein.
\newblock Can you learn an algorithm? generalizing from easy to hard problems
  with recurrent networks.
\newblock \emph{Advances in Neural Information Processing Systems}, 34,
  2021{\natexlab{b}}.

\bibitem[Selsam et~al.(2018)Selsam, Lamm, B{\"u}nz, Liang, de~Moura, and
  Dill]{selsam2018learning}
Daniel Selsam, Matthew Lamm, Benedikt B{\"u}nz, Percy Liang, Leonardo de~Moura,
  and David~L Dill.
\newblock Learning a sat solver from single-bit supervision.
\newblock \emph{arXiv preprint arXiv:1802.03685}, 2018.

\bibitem[Slotine et~al.(1991)Slotine, Li, et~al.]{slotine1991applied}
Jean-Jacques~E Slotine, Weiping Li, et~al.
\newblock \emph{Applied nonlinear control}, volume 199.
\newblock Prentice hall Englewood Cliffs, NJ, 1991.

\bibitem[Soldaini and Moschitti(2020)]{DBLP:journals/corr/abs-2005-02534}
Luca Soldaini and Alessandro Moschitti.
\newblock The cascade transformer: an application for efficient answer sentence
  selection.
\newblock \emph{CoRR}, abs/2005.02534, 2020.
\newblock URL \url{https://arxiv.org/abs/2005.02534}.

\bibitem[Sriperumbudur and Lanckriet(2009)]{sriperumbudur2009convergence}
Bharath~K Sriperumbudur and Gert~RG Lanckriet.
\newblock On the convergence of the concave-convex procedure.
\newblock In \emph{Nips}, volume~9, pages 1759--1767. Citeseer, 2009.

\bibitem[Teerapittayanon et~al.(2016)Teerapittayanon, McDanel, and
  Kung]{teerapittayanon2016branchynet}
Surat Teerapittayanon, Bradley McDanel, and Hsiang-Tsung Kung.
\newblock Branchynet: Fast inference via early exiting from deep neural
  networks.
\newblock In \emph{2016 23rd International Conference on Pattern Recognition
  (ICPR)}, pages 2464--2469. IEEE, 2016.

\bibitem[Vaswani et~al.(2017)Vaswani, Shazeer, Parmar, Uszkoreit, Jones, Gomez,
  Kaiser, and Polosukhin]{vaswani2017attention}
Ashish Vaswani, Noam Shazeer, Niki Parmar, Jakob Uszkoreit, Llion Jones,
  Aidan~N Gomez, {\L}ukasz Kaiser, and Illia Polosukhin.
\newblock Attention is all you need.
\newblock In \emph{Neural Information Processing Systems}, 2017.

\bibitem[Wang et~al.(2003)Wang, Lee, and Lin]{wang2003global}
Qing-Guo Wang, Tong~Heng Lee, and Chong Lin.
\newblock Global stability of limit cycles.
\newblock In \emph{Relay Feedback}, pages 57--83. Springer, 2003.

\bibitem[Wang et~al.(2020)Wang, Zhang, and Sun]{wang2020implicit}
Tiancai Wang, Xiangyu Zhang, and Jian Sun.
\newblock Implicit feature pyramid network for object detection.
\newblock \emph{arXiv preprint arXiv:2012.13563}, 2020.

\bibitem[Wang et~al.(2018)Wang, Yu, Dou, Darrell, and
  Gonzalez]{wang2018skipnet}
Xin Wang, Fisher Yu, Zi-Yi Dou, Trevor Darrell, and Joseph~E Gonzalez.
\newblock Skipnet: Learning dynamic routing in convolutional networks.
\newblock In \emph{Proceedings of the European Conference on Computer Vision
  (ECCV)}, pages 409--424, 2018.

\bibitem[Winston and Kolter(2020)]{winston2020monotone}
Ezra Winston and J~Zico Kolter.
\newblock Monotone operator equilibrium networks.
\newblock \emph{arXiv preprint arXiv:2006.08591}, 2020.

\bibitem[Wu and He(2018)]{wu2018group}
Yuxin Wu and Kaiming He.
\newblock Group normalization.
\newblock In \emph{Proceedings of the European conference on computer vision
  (ECCV)}, pages 3--19, 2018.

\bibitem[Xing et~al.(2020)Xing, Xu, Li, and Guan]{xing2020early}
Qunliang Xing, Mai Xu, Tianyi Li, and Zhenyu Guan.
\newblock Early exit or not: Resource-efficient blind quality enhancement for
  compressed images.
\newblock In \emph{European Conference on Computer Vision}, pages 275--292.
  Springer, 2020.

\bibitem[Zhou et~al.(2020)Zhou, Xu, Ge, McAuley, Xu, and Wei]{zhou2020bert}
Wangchunshu Zhou, Canwen Xu, Tao Ge, Julian McAuley, Ke~Xu, and Furu Wei.
\newblock Bert loses patience: Fast and robust inference with early exit.
\newblock \emph{Advances in Neural Information Processing Systems},
  33:\penalty0 18330--18341, 2020.

\end{thebibliography}

\appendix

\newpage
\section{Alternative Approaches for Quantifying Path Independence}
\label{sec:alternative_pi_metric}
We review alternative methods for quantifying how path-independent a given equilibrium model is, and why the Fixed Point Alignment score is the most suitable one amongst them for our analyses. We want a path independence metric to satisfy three criteria: 1) \textbf{Dimensionless:} Metrics computed from different training runs should be directly comparable with each other. To ensure this, one has to make sure that one uses dimensionless metrics (i.e. doesn't have units). 2) \textbf{local and global path independence:} The metric should test for path independence not just in a local neighborhood of fixed points, but over a sufficiently large portion of the initialization domain. 3) \textbf{efficiency:} It should be computationally tractable. 

With these in mind, we review three alternative ways of quantifying path independence. The summary of the analysis is described in Table :
\begin{itemize}
    \item \textbf{Jacobian Norm:} The Frobenius norm of the Jacobian of the output of an equilibrium model with respect to its fixed point initialization (i.e. $||\pp{\mathrm{FIX}(\vx, \vz_0)}{\vz_0}||_2$) gives a robust measure of how locally sensitive the network is to initializations. If this quantity is very small (and approaching $0$ with more root finding steps), one can conclude that the given equilibrium model is path-independent on the given input. The main issues with this approach are: 1) It's extremely computationally intensive to compute (since both the input and the output of $\mathrm{FIX}$ are high-dimensional, neither forward nor backward differentiation can estimate this Jacobian efficiently) 2) It has units, meaning it isn't comparable across different training runs, 3) It only measures local path independence. 

    \item \textbf{Agreement with Adversarial Initializations: } This method quantifies to what extent it is possible to directly optimize initializations such that they result in different fixed points. If the network is path independent, there shouldn't exist an adversary that can find adversarial initializations. The exact procedure for how adversarial fixed points can be found in Section \ref{sec:stres_test_appendix}. The downsides with this approach is that 1) it's expensive - one has to solve an optimization problem for every problem instance, potentially using different optimizers and initializations, and 2) it only checks for local path independence. 

    \item \textbf{Agreement with Swapped Initializations (Fixed Point Alignment): } The main idea behind this approach is to initialize the forward pass of an equilibrium model with fixed points obtained from other problem instances, and checking if the network still displays the same asymptotic behaviour (i.e. finds the same fixed point). If one uses cosine similarity to measure consistency, then one recovers Fixed Point Alignment score proposed in Section \ref{sec:def_fpa}. Another option is to use the average Euclidean distance to check for consistency. This is identical to computing the trace of the covariance matrix of the distance between the computed fixed points. 
    This approach -- especially the one that utilizes cosine similarity to quantify consistency -- is appealing, as 1) because it's unitless, it can be compared between training runs on the same task, 2) it checks for non-local convergence by sampling from a distribution of relevant fixed points 3) it's very efficient, requiring only two forward passes. The version that utilizes Euclidean distance does have units, hence cannot be used for cross-model comparisons~\footnote{If one uses LayerNorm~\citep{ba2016layer} or similar normalization layers, we could expect the Euclidean distance based metric to behave similar to the cosine similarity based one.})
    To further make sure that this metric is correct, and does subsume with local path independence, we ran stress tests described in Section \ref{sec:def_fpa} to make sure this metric produces results consistent with the \textit{agreement with adversarial initializations} method described above. 
\end{itemize}

\paragraph{Alternatives to Cosine Similarity in Measuring Directional Alignment}
While cosine similarity is a conceptually clean way of quantifying directional alignment, one can consider alternative kernels as well. Ideally, our main results should not depend on the specific implementation details of path independence metrics (as long as they satisfy the criteria we set out above). 

\begin{figure}
  \begin{center}
 \includegraphics[width=.99\textwidth]{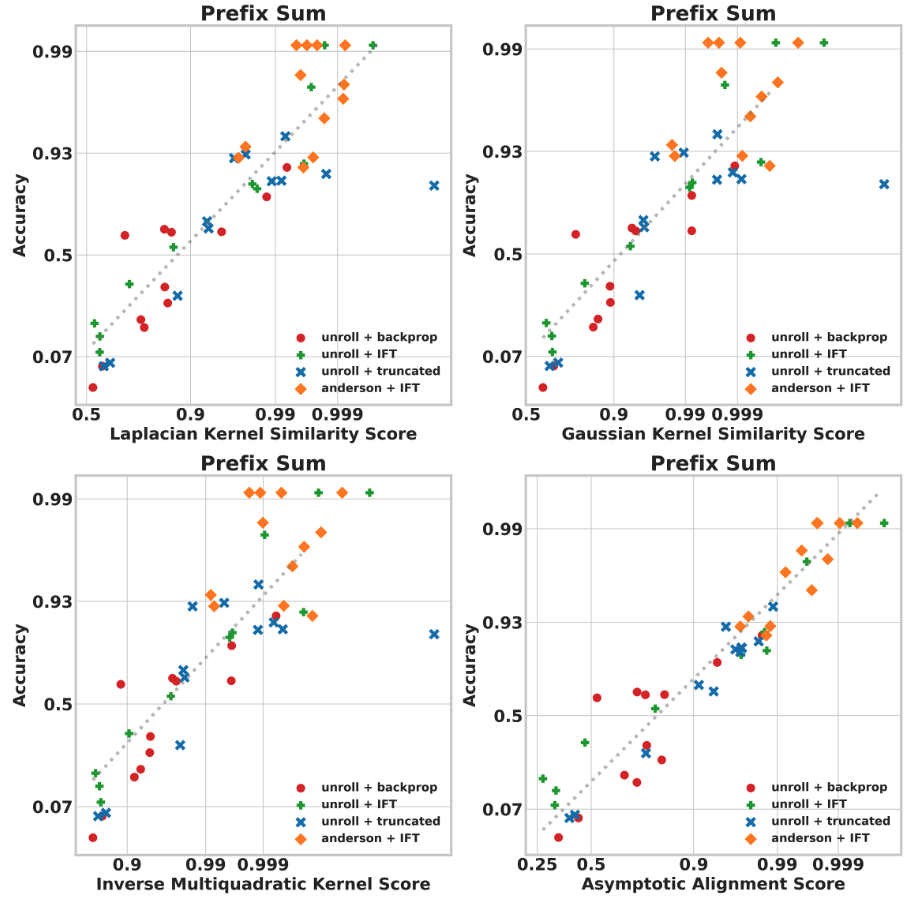}
  \end{center}
    \caption{\textbf{Alternatives to Cosine Similarity in AA Score:} Replacing cosine similarity (right bottom) with alternative kernels like Gaussian (left top), Laplacian (right top) and inverse multiquadratic kernel yields qualitatively similar results. This demonstrates that the takeaways from our experiments don't depend on the particular implementation details of the ways path-independence is quantified. } 
    \label{fig:other_kernels} 
    \vspace{-3mm}
\end{figure}

To check how sensitive our results are on the precise functional form of the AA score, we replaced it with three other kernels and re-assessed whether the results pointed to the same high-level takeaways. Concretely, we tried the Gaussian kernel $\phi_g(r) = \exp (-(\eps r)^2)$, Laplacian kernel $\phi_l(r) = \exp (-|\eps r|)$ and inverse multiquadratic kernel $\phi_i(r) = \frac{1}{ \sqrt{1 + (\eps r)^2}}$ where the input to the kernel function is the Euclidean distance of the normalized fixed points $r = || \frac{z_1}{||z_1||} - \frac{z_2}{|| z_2 ||} ||$. We used $\eps = 5000$, which gave a reasonable dynamic range in the resulting values. The results can be seen in Figure \ref {fig:other_kernels}. The takeaways from the plots remain identical: \textit{path independence is strongly correlated with generalization, regardless of the specific details of how path independence is quantified} (as long as it satisfies the criteria we set in Appendix A). 

\textbf{Importance of using dimensionless metrics: } Being unitless is necessary for a metric to be comparable across training runs, though obviously not sufficient. Consider two equilibrium models M1 and M2, where the fixed points computed by M2 have the same direction as those computed by M1, but have twice the Euclidean norm. The behaviour of this M2 is qualitatively the same as that of M1, but any metric that depends on the Euclidean metric (or Jacobian L2 norm, or any non-unitless metric) would report this network to be less path independent. Note that this is not a purely theoretical consideration: simply adding L2 regularization, or penalizing the magnitude of fixed points to encourage convergence will directly impact the scale of the fixed points (things that practitioners often use), thereby rendering non-unitless metrics unreliable.

\begin{table*}[t]
\resizebox{\textwidth}{!}{%
\begin{tabular}{l|ccc}
\toprule
\textbf{Path Independence Metrics}  &  \textbf{Dimensionless} & \textbf{Local \& Global Coverage} & \textbf{Efficient}  \\ \midrule
IO Jacobian Norm & \efficacylow & \efficacylow & \efficacylow \\
Using Adversarial Initializations & \efficacyhigh & \efficacymed & \efficacylow \\
Fixed Point Alignment Score  & \efficacyhigh & \efficacymed & \efficacyhigh \\
\bottomrule
\end{tabular}}
\caption{\textbf{Comparing different ways of quantifying path independence:} We compare three different methods of quantifying how path independent a network is in terms of their correctness (whether they actually measure path independence), dimensionlessness (whether the quantity is unitless, allowing comparisons between between different networks meaningful), local and global coverage (whether the metric checks for path independence locally or globally) and efficiency (whether it's computationally cheap to compute the metric). Among the methods we've considered (See Section \ref{sec:alternative_pi_metric}), the Fixed Point Alignment Score is the most suited one for our purposes. Note that no metric considered here has perfect global coverage (i.e. can verify that a given network is globally path independent or not). }
\vspace{-0.5cm}
\label{table:lg_options}
\end{table*}

\section{Experimental Details}
This section contains the additional information needed to reproduce our experimental findings. This section complements our code release. 
\subsection{Details for Experiments on Prefix Sum}
\label{app:prefix_training_details}
Unless otherwise specified, all of the prefix-sum networks were trained under the following conditions: 

\textbf{Training details:} We used the Adam~\citep{kingma2014adam} optimizer and trained for $30000$ gradient steps with a batch size of $150$. We varied the initial learning rate on different runs ($0.001$, $0.0001$ and $0.00001$). We applied a stepwise learning rate decay both with a factor of $0.5$ halfway and three-quarters through the learning. We also used gradient clipping at L2 norm equals 1. 

\textbf{Model selection:} Since we didn't observe overfitting in our experiments, we used the last model saved model checkpoint during evaluation. 

\textbf{Training data:} We used the length-32 training split of the prefix sum dataset described in \citet{schwarzschild2021datasets}. 

\textbf{Architecture:} We used a 1D-convolutional depth-wise recurrent ResNet architecture roughly analogous to that used by \citet{bansal2022end} in their prefix sum experiments. The main differences include: (1) We used a single convolutional readout layer, as well as single convolutional layer to right before the depth-wise recurrent section of the network.  (2) Unless otherwise specified, we initialized the fixed point at the $0$ vector, unlike \citet{bansal2022end} who initialized the fixed point vector directly with the input. (3) While \citet{bansal2022end} uses concatenation to do input injection, we directly added the input (or a projection of it) to the hidden activations of the residual blocks, the same way the \citet{kolter2020tutorial} does in their convolutional architecture. (4) We ablated using Weight Normalization~\citep{DBLP:journals/corr/SalimansK16}, as this is something that has been reported to stabilize training. 

\textbf{Forward solvers:} We used either fixed point iterations (of depth $6$ or $32$) or Anderson acceleration~\citep{kolter2020tutorial} (with an iteration count of $6$ or $32$ and a memory of $3$) depending on the experimental purpose of the experiment. 

\textbf{Backwards pass:} We used either the standard backpropagation algorithm, the IFT gradients to train the networks or truncated backpropagation, depending on the purpose of the experiment. We used Anderson Acceleration to estimate the IFT gradients with an iteration budget of $10$ and memory of $3$. We found that rescaling the Jacobian term that appears in Equation \ref{eq:deq-backward-pass}  with a constant less than $1$  stabilized training with the IFT gradients. We used a rescaling factor of $0.8$. While this is relatively non-standard practice, we confirmed that using this stabilization strategy doesn't yield qualitatively different results than those reported in the paper. When we used truncated backpropagation, we dropped the gradients halfwah through the network (i.e. if the forward pass consisted of 10 iterations of the recurrent cell, we only backpropagated through the last 5 iterations). 

\textbf{Loss function: } We used the cross entropy between the network predictions and the labels as the loss function. the prefix-sum task is a sequence-to-sequence problem, and each forward pass consists of multiple predictions. We used the average of all of the cross entropy values computed in a single forward pass. 

\textbf{Compute resources:} All of the prefix sum networks were trained using Intel Xeon Silver CPUS and NVIDIA GPUs (both P100, T4V1 and T4V2 models were used in the experiments). Training one prefix-sum network took roughly $1$ to $2$ hours depending on how deep the forward pass is, and whether one is using IFT or backpropagation gradients. 

\subsubsection{Reproducing Table \ref{table:adversarial-fpa-mazes-prefix-sum} (Stress-Testing the AA Score)}

The non-path independent network used in this table is a $15$-layer weight-tied input-injected network trained with backpropagation gradients. We set the maximum number of iterations for fixed point computation to $11$ as we found that this choice gives the highest test accuracy on 64 bit prefix sum strings. While selecting the optimal value for maximum iterations, we increased the value up to $55$ iterations. The path independent network is a $32$-layer weight-tied input inejcted network trained with backpropagation gradients. The maximum iterations for this network are set to $500$. We used a  batch size of $1$ and set the maximum number of L-BFGS \citep{liu1989limited} updates per batch to $50$ for both the networks. We implemented this code in PyTorch and use these values of hyperparameters: \texttt{lr=1}, \texttt{tolerance\_grad}=1e-7, \texttt{tolerance\_change}=1e-9, \texttt{line\_search\_fn="strong\_wolfe"}. We report AA scores and accuracy computed on $500$ examples. We provide pseudocode in Algorithm Box \ref{alg:adveresarial-lbfgs}

\subsection{Details for Experiments on Mazes}
\label{app:maze_training_details}
Unless otherwise specified, all the networks for mazes were trained under the following conditions: 

\textbf{Training details:} We used the Adam~\citep{kingma2014adam} optimizer with an  initial learning rate of $0.001$ and batch size of $50$. We use stepwise learning rate decay with a factor of $0.1$ applied at epoch number $100$. We trained all weight-tied input-injected networks for a total of 150 epochs. We observed that implicitly trained equilibrium models needed a longer training time for their validation accuracy to improve. All implicitly trained equilibrium models were pretrained for $150$ epochs and trained with IFT for another $100$ epochs. 

\textbf{Model selection:} We used the checkpoint corresponding to the model with the best validation accuracy during evaluation. If multiple checkpoints had same validation accuracy, we selected the most recent checkpoint.

\textbf{Training data:} We trained on the training split of $9\times9$ mazes from the dataset described in \citet{schwarzschild2021datasets}. 

\textbf{Architecture:} We used a 2D-convolutional depth-wise recurrent ResNet architecture analogous to that used by \citet{bansal2022end} in their mazes  experiments. While pretraining implicitly trained equilibrium models, we do rollouts for $32$-iterations. For equilibrium models trained with phantom gradient in Table \ref{table:training-time-convergence}, we use a damping factor of $0.75$ and iterate for 2 iterations. Shallow weight-tied input-injected networks were trained with $8$-layers while the deeper weight-tied input-injected networks used $32$ layers. Both the feedforward networks are $32$-layer deep non-weight tied networks. We ablate on input-injection for these networks. We used the original code released by \citep{bansal2022end} to replicate the results for weight-tied input-injected networks trained with progressive training. In addition we ablate on weight normalization~\citep{DBLP:journals/corr/SalimansK16} for all the architectures---both unrolled and implicitly trained equilibrium models. We found that weight normalization helps stabilize the training, especially implicitly trained models. We found that using normalization layers (like group norm \citep{wu2018group}) hurts generalization in mazes unlike prefix sum. Therefore, none of our architectures for mazes use any normalization layers.

\textbf{Forward solvers:} We used either fixed point iterations (of depth $8$ or $32$) or Broyden's method~\citep{broyden1965class} (with an iteration count of $40$) depending on the experimental purpose of the experiment. 

\textbf{Backward pass:} We used the standard backpropagation algorithm, the IFT gradients, truncated backpropagation or phantom gradients to train the networks, depending on the purpose of the experiment. We used Broyden's method to estimate the IFT gradients with an iteration budget of $40$. For experiments with truncated backpropagation, we follow the exact setting as specified in \citep{bansal2022end}.

\textbf{Loss function: } We used the cross entropy between the network predictions and the labels as the loss function. 

\textbf{Compute resources:} All networks for mazes were trained using Intel Xeon Silver  CPUs and NVIDIA GPUs (1080Ti, 2080Ti and A6000 models were used in the experiments). Training a weight-tied network took roughly around $12$-$18$ hours depending on how deep the forward pass is, whereas the implicitly trained models took around $2$ days of training time on a single GPU. 

\subsection{Reproducing Figure~\ref{fig:trajectory_plot}}

Both models were trained using 32 fixed point iterations and backpropagation. The path independent model is input injected, the path dependent model is not input injected. We run the network for 32 fixed point iterations on five different validation examples of length 32 to obtain different hidden layer initializations. We then run the models for 128 steps using these initializations to obtain five different trajectories. For the input injected network, we use the same input injection for all trajectories, obtained from a different example of length 32. We project the resulting hidden states onto two random orthogonal directions in the hidden layer state space.

\subsection{Experimental Details for Section \ref{sec:algo_ood_eq} (\nameref{sec:algo_ood_eq})}
\label{sec:stres_test_appendix}
\subsubsection{Reproducing Figure \ref{fig:ood_generalization_mazes} (OOD Generalization with Mazes)}
The training details for all the weight-tied input-injected networks used to produce this plot can be found in Section \ref{app:maze_training_details}. The feedforward network used in this plot is a $32$-layer non-weight tied network without any input injection that was trained with regular backpropagation. The maximum number of test-time iterations (both unroll and Broyden solver) for mazes of size $9$, $13$, $25$, $31$ and $59$ were $50$, $100$, $400$, $600$, and $800$ respectively. The test accuracy for mazes of a given dimension was computed on the entire split of $10,000$ mazes as provided in the original dataset \citep{schwarzschild2021can}.

\subsubsection{Reproducing Figure \ref{fig:pi-vs-nonpi-accuracy} (PI networks get better with more iterations)}

The training details for the $12$ networks used to produce this plot can be found in Section \ref{app:prefix_training_details}. We varied (1) the depth ($6$ vs. $32$), whether weight normalization\citep{DBLP:journals/corr/SalimansK16} was used or not, and the initial learning rate ($0.001$, $0.0001$ or $0.00001$).

We evaluated each model on 128 examples of length 16, 32, 64, and 128 each, using fixed point iterations and measuring the average number of correct bits at each iteration. For each model and iteration step, we plot average accuracies over all splits and examples. The Fixed Point Alignment Scores for each model are calculated as in Figure~\ref{fig:pi-corr-prefix-sum} (see Appendix~\ref{pi-accuracy-figure-description-prefix-sum}).

\subsubsection{Reproducing Table \ref{table:adversarial-fpa-mazes-prefix-sum} (Stress-Testing the AA Score)}

The non-path independent network used in this table is a $8$-layer weight-tied input-injected network trained with backpropagation gradients. We set the maximum number of iterations for fixed point computation to $12$ as we found that this choice gives the highest test accuracy on $13\times13$ mazes. While selecting the optimal value for maximum iterations, we increased the value up to $60$ iterations. The path independent network is a $32$-layer weight-tied input injected network trained with backpropagation gradients. The maximum iterations for this network are set to $500$. We used a  batch size of $1$ and set the maximum number of L-BFGS \citep{liu1989limited} updates per batch to $50$ for both the networks. We implemented this code in PyTorch and use these values of hyperparameters: \texttt{lr=1}, \texttt{tolerance\_grad}=1e-7, \texttt{tolerance\_change}=1e-9, \texttt{line\_search\_fn="strong\_wolfe"}. We report AA scores and accuracy computed on $500$ examples. We provide pseudocode in Algorithm Box \ref{alg:adveresarial-lbfgs}
\begin{algorithm}[H]
    \centering
    \caption{Stress testing AA Scores: An algorithm to find adversarial initializations } \label{alg:naive-algorithm}
    \begin{algorithmic}
        \Require A trained network $f_\vw$, an input $\vx$, $N$ max number of LBFGS updates\\
        \textbf{Define:}  $h(\vy_1, \vy_2) := \dfrac{\vy_1}{\| \vy_1 \|_2} \cdot \dfrac{\vy_2}{\|\vy_2 \|_2}$\\
        random init($\cdot$): A method that returns a random vector initialized from  $\mathcal{N}(\mathbf{0}, \mathbf{I})$ \\
        \textbf{Initialize:} $\vz = \vx$ or random init($\vx$)
        \LeftComment Disable gradients
        \State $\vz_1$ = $\mathrm{FIX}_{f_\vw}(\vz, \vx)$
        \LeftComment Clone $\vz_1$ and Enable gradients on $\vz_1$
        \For{trials from $1$ to $N$}
            \State $\vz_2$ = $\mathrm{FIX}_{f_\vw}(\vz_1, \vx)$ 
            \State Perform L-BFGS update on $\vz_1$ to minimize 
                $h(\vz_1, \vz_2)$
        \EndFor
        \Ensure $\vz_1$
    \end{algorithmic}\label{alg:adveresarial-lbfgs}
\end{algorithm}

\subsection{Experimental Details for Section \ref{sec:pi-vs-ood} (\nameref{sec:pi-vs-ood})}

\subsubsection{Reproducing Figure \ref{fig:pi-corr-prefix-sum} (Path Independence - Accuracy Correlation on Prefix Sum)}
\label{pi-accuracy-figure-description-prefix-sum}
The training details for the $12$ networks used to produce this plot can be found in Section \ref{app:prefix_training_details}. We trained $12$ networks under each category reported in the plot, where (1) the depth ($6$ vs. $32$), whether weight normalization\citep{DBLP:journals/corr/SalimansK16} was used or not, and the initial learning rate ($0.001$, $0.0001$ or $0.00001$) were varied. 

Each value on the plots were computed using $300$ examples from the validation splits corresponding to lengths $16$, $32$, $64$, $128$ and $256$ (i.e. $5*300 = 1500$ samples per point). The average cosine similarities involved in computing the Fixed Point Alignment scores were computed using $10$ different initialization per example. For the test-time forward pass, we used $512$ iterations of Anderson Acceleration~\citep{kolter2020tutorial} (note that this is using a more sophisticated solver using $16$ times more forward iterations than the deepest of the trained networks). The trend lines on the plots were computed using the line that minimizes the least squares error.

\subsubsection{Reproducing Figure \ref{fig:pi-corr-mazes} (Path Independence - Accuracy Correlation on Mazes)}
The architectural and training details for all the networks in this plot can be found in Section \ref{app:maze_training_details}. Each value in this plot was computed using $500$ examples from the validation splot corresponding to mazes of size 13 and 25. The average cosine similarities involved in computing the Fixed Point Alignment scores were computed using $10$ different initialization per example. Unrolled + backprop networks use $100$ test-time iterations for the forward pass for mazes of size $13\times13$, and $400$ test-time iterations for mazes of size $25\times25$.  We set the maximum number of permissible iterations for the Broyden solver \citep{broyden1965class}, and for non-weight tied feedforward network to the same values. The shallow unrolled + backprop network was trained using $8$ layers. Empirically, we obtained the highest test accuracies on $11^{th}$ and $14^{th}$ iterations for mazes of size $13\times13$ and $25\times25$ respectively, for networks that use weight normalization. For networks that do not use weight normalization, the corresponding numbers are reported on $12^{th}$ and $18^{th}$ iteration. The trend lines on the plots were computed using the line that minimizes the least squares error. 

\subsection{Experimental Details for Section \ref{sec:manipulations} (\nameref{sec:manipulations})}

\subsubsection{Reproducing Figure \ref{mix:algo_and_intervention} (Manipulations of PI)}
The training details for the $12$ networks that formed the baseline for the interventions are described in Section \ref{app:prefix_training_details}. We trained $12$ networks under each category reported in the plot, where (1) the depth ($6$ vs. $32$), whether weight normalization\citep{DBLP:journals/corr/SalimansK16} was used or not, and the initial learning rate ($0.001$, $0.0001$ or $0.00001$) were varied. 

For each intervention, we kept all of the aforementioned training conditions except for the experimental manipulations that are intended to promote or hurt path independence. 
\begin{itemize}
    \item \textbf{Mixed initialization:} During each training forward pass, each sample was assigned with either zero initialization (i.e. the fixed point was initialized with the $0$ vector) or standard normal distribution (i.e. ...) using a Bernoulli random variable of probability $0.5$ (i.e. the examples that were run with zero vs. normal initializations were roughly half-half. 
    \item \textbf{Randomized depth:} Randomized training was done by first specifying the minimum and maximum possible depths, then uniformly sampling a depth from that range during each forward pass. We used a range of $3$ to $63$ forward iterations for training $6$ of the networks, and used a range of $32$ to $64$ for training the other $6$. 
    
    \item \textbf{Alignment penalty: } The alignment penalty, which was added to the original training loss in form of a regularized was computed as follows: (1) For each example in the batch, compute the fixed points $k$ times using normal initialization, (2) Compute the dot product between each pair of fixed points (which yields $k^2 - k$ dot products, excluding each fixed point's dot product with itself), return the sum of the dot products, normalized by $k^2 - k$. In our experiments, we picked $k=3$. 
\end{itemize}

\subsection{Experimental Details for Section \ref{sec:disambiguate-convergence-pi} (\nameref{sec:disambiguate-convergence-pi})}

\subsubsection{Reproducing Table \ref{table:training-time-convergence} (Training Convergence vs. Accuracy)}
The architectural and training details for equilibrium models trained with phantom gradients and with IFT have been provided in Section \ref{app:maze_training_details}. Both the networks use Broyden solver for the forward pass. We set the number of maximum permissible iterations to $40$ for in-distribution examples  (\ie $9\times9$ mazes) and to $400$ for $25\times25$ mazes. All the values reported in this table have been computed on $500$ examples from the respective splits for in-distribution and out-of-distribution data. 

\subsubsection{Reproducing Figure \ref{fig:test-time-residuals} (Test Time Convergence vs. Path-Independence)}
The architectural and training details of weight-tied input-injected networks used in this figure have been provided in Section \ref{app:maze_training_details}. The reported numbers have been averaged over 10 different runs. Each run uses a batch size of 100. The maximum number of forward iterations is set to 200 for both fixed-point and Broyden solvers. 

\subsubsection{Reproducing Figure \ref{fig:pis-corr-incorr-ver} (Per Instance Path Independence vs. Accuracy)}
To compute the histogram, we took the $12$ networks trained with the training condition that yielded the strongest OOD accuracy (mixed initialization) and computed both the accuracy and AA scores on $300$ examples for the prefix-sum dataset validation splits of lengths $16$, $32$, $64$, $128$ and $256$. We grouped each AA score based on whether the prediction it's paired with was correct or not, then produced the histogram.

\section{Limitations}
Currently, the experimental results are obtained only using two tasks: prefix-sum and mazes. Testing the validity of our findings on a more diverse set of tasks would increase our confidence in the fact that the phenomenon reported in this paper is universal. Moreover, it is not clear from our analysis which tasks would benefit from more test-time compute. While the necessity of additional test-time compute is obvious for prefix sum and mazes, it's less clear for perception-like tasks like image classification. Furthermore, we believe that a more systematic study of how the choice of solver (i.e. Anderson (first order) vs. Broyden (second order) solvers) impacts the empirical AA scores would strengthen our findings. We observe, for example, that networks trained with Broyden struggle to find fixed points at test time when run with fixed-point iterations. Expanding on these findings would improve our analysis and let us make more concrete suggestions to practitioners aiming to use the path-independence related concepts from this paper.

\section{Path Independence vs. Accuracy Plots Different Difficulty Levels}
\label{sec:pi_vs_acc_on_diff_len}
\begin{figure}
  \begin{center}
 \includegraphics[width=1.\textwidth]{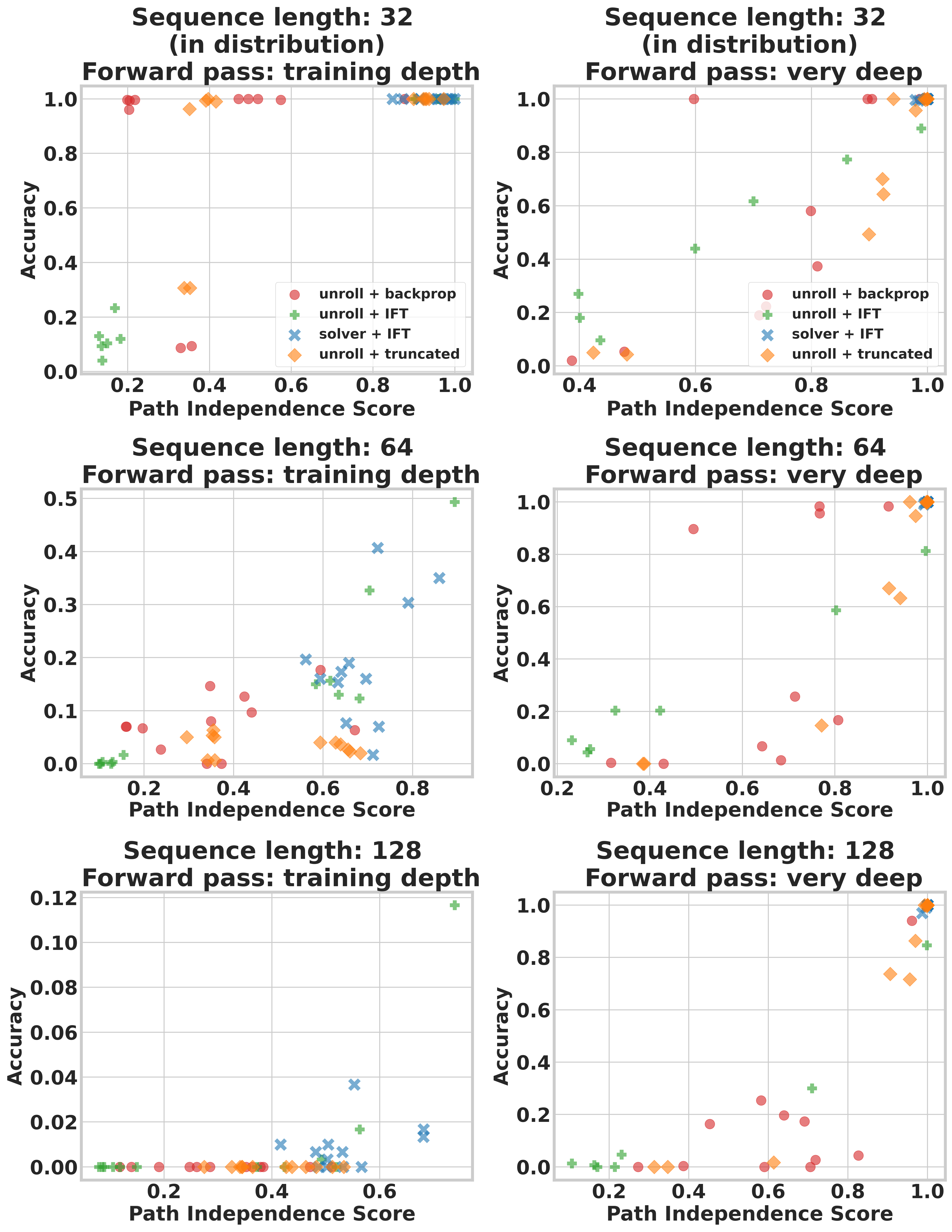}
  \end{center}
    \caption{\textbf{Accuracy vs. Path Independence Correlation on Different Problem Difficulties and Inference Depths on Prefix Sum:} The data on the plots in the left column is obtained when the forward pass is run with the training forward pass budged. Likewise, the data on the right column is obtained in the large inference-time budget limit. Each row corresponds to a different problem difficulty, with the first row (length = 32) corresponding to in-distribution data. } 
    \label{fig:pi_vs_acc_different_lens} 
\end{figure}

The path-independence vs. accuracy plots in Figure \ref{fig:pi-corr-mazes} uses averaged values over different problem difficulties. In Figure \ref{fig:pi_vs_acc_different_lens}, we show how these correlations look like at different lengths on the prefix sum task. The same figure also shows how the accuracy differs if one does inference using the training time forward pass budget. The main takeaways from this plot are:
\begin{itemize}
    \item Out-of-distribution problem lengths (right top of the Figure) do indeed require larger inference depths - all models perform very poorly when inference is run with training-time forward pass budget. 
    \item The correlation between path independence and accuracy - as the definition implies - is more apparent in the very large forward pass limit. In this regime, the correlation persists for all lengths. 
    \item While networks with low PI values using the training-time budget occasionally perform worse when the test-time compute is increased, that doesn't happen with path-independent networks. 
\end{itemize}

\section{Intervention Results on Different Difficulty Levels}
\begin{figure}
  \begin{center}
 \includegraphics[width=1.\textwidth]{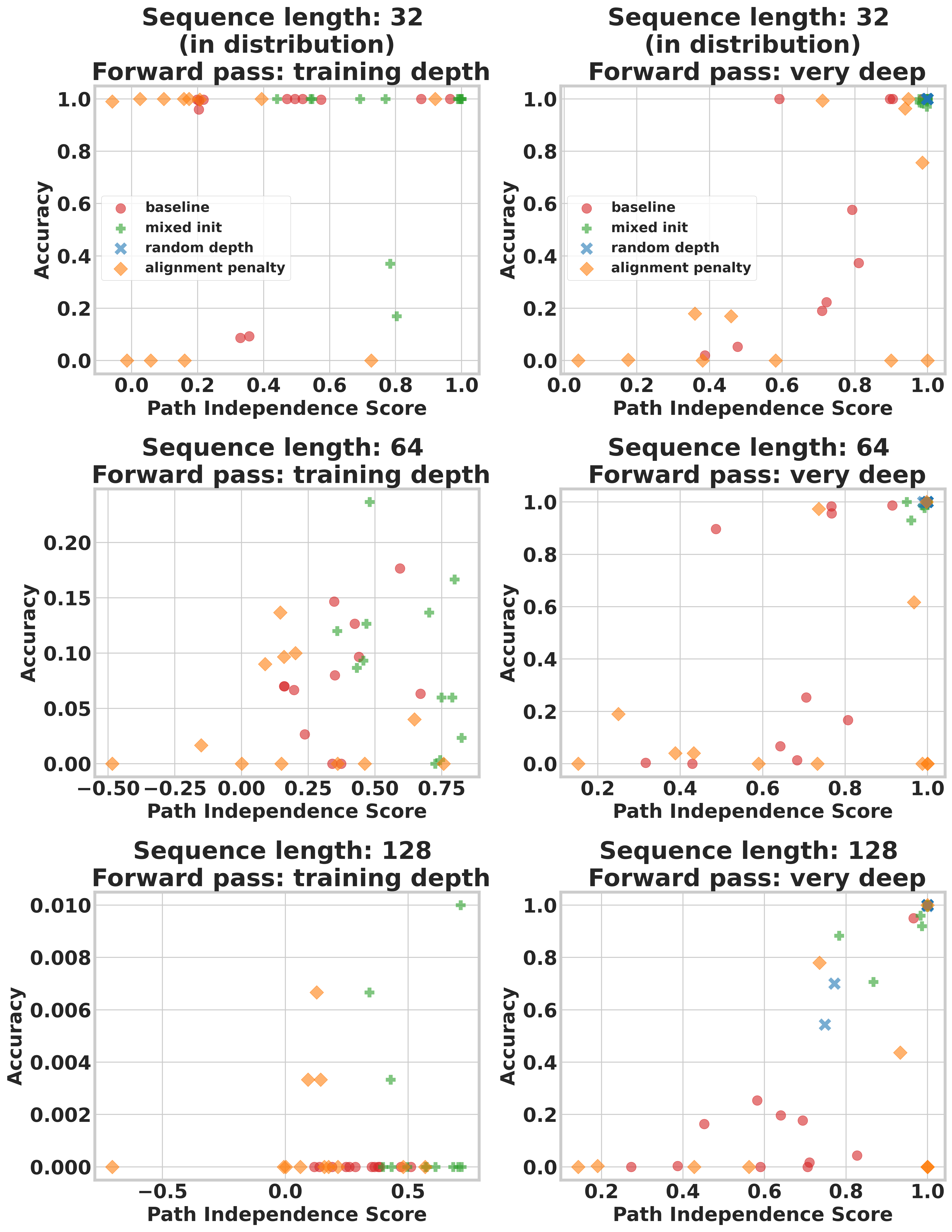}
  \end{center}
    \caption{\textbf{Accuracy vs. Path Independence Correlation on Different Problem Difficulties and Inference Depths on Prefix Sum:} The data on the plots in the left column is obtained when the forward pass is run with the training forward pass budged. Likewise, the data on the right column is obtained in the large inference-time budget limit. Each row corresponds to a different problem difficulty, with the first row (length = 32) corresponding to in-distribution data. } 
    \label{fig:pi_vs_acc_intervention_different_lens} 
\end{figure}

Mirroring Section \ref{sec:pi_vs_acc_on_diff_len}, we provide how the accuracy correlates with path independence on 1) different problem difficulties ($32$ is in-distribution) and different forward pass budgets in Figure \ref{fig:pi_vs_acc_intervention_different_lens}.  

We especially emphasize the \textit{mixed initialization} results (mixed initialization is one of the interventions that promotes path independence). Note that some of these networks (half of which are as shallow as $6$ layers) do poorly in and out-of-distribution when the forward pass uses the training  forward pass budget. Unsurprisingly, the AA scores for these networks are low in this condition. However, when additional test-time compute is provided, these networks reach perfect in-distribution accuracy, and near-perfect OOD accuracy (especially on length = 64 split). This demonstrates that training time convergence is not required for path independence. We find it surprising that with the help of a path-independence-promoting regularization procedure, it's possible to train very shallow (i.e. less than $6$ layers) networks such that they can exploit additional test-time compute to achieve very high accuracy.

\section{Per-Instance Path Independence Analyses - Convergence vs. Path Independence}

\begin{figure}
  \begin{center}
 \includegraphics[width=.8\textwidth]{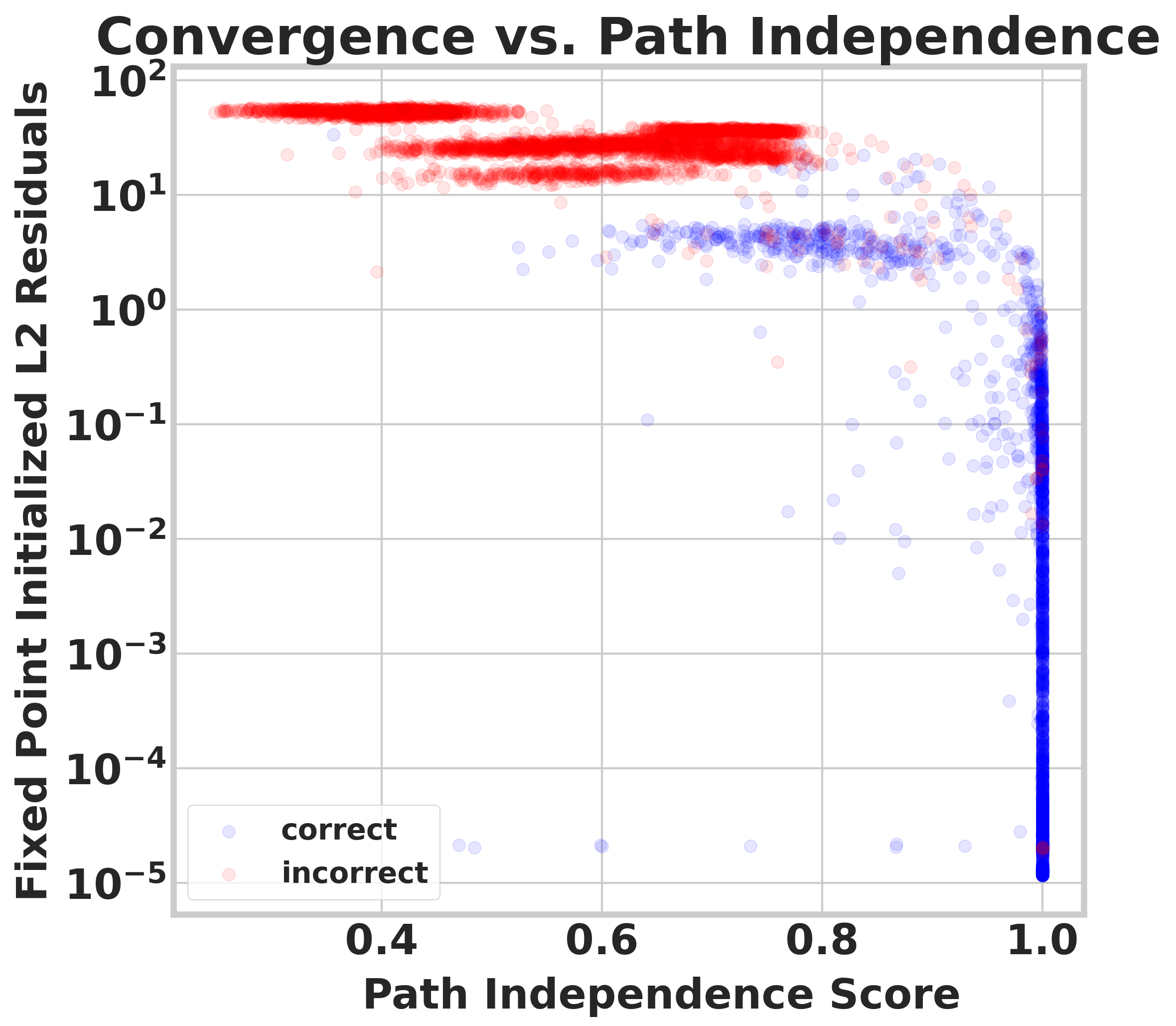}
  \end{center}
    \caption{\textbf{Per-Instance Test-time Convergence vs. Path Independence}: The per-instance AA score vs. convergence (measured in terms of the L2 distance between the last two root solver iterates). Each example is labelled based on whether the network's prediction on it was correct (blue) or not (red). We used the $12$ networks trained using the mixed initialization strategy (since these networks yielded the best in and OOD performance), and used $300$ samples from the length splits $16$, $32$, $64$, $128$ and $256$. While good convergence is almost always associated with high AA score values, the converse is not necessarily true: there are many samples on which convergence is poor, yet the AA score is high. } 
    \label{fig:conv_vs_pi_analysis} 
\end{figure}

In addition to the analysis in Section \ref{sec:per_instance_analysis}, we also analyzed how test-time convergence correlates with path independence on a per-instance basis. The per-instance AA score vs. convergence (measured in terms of the L2 distance between the last two root solver iterates) can be seen in Figure \ref{fig:conv_vs_pi_analysis}. We used the $12$ networks trained using the mixed initialization strategy (since these networks yielded the best in and OOD performance), and used $300$ samples from the length splits $16$, $32$, $64$, $128$ and $256$. 

This analysis reaffirms our past observation that test-time convergence is not needed for path independence. The plot contains data from three different regimes:
\begin{itemize}
    \item \textbf{Full convergence to a fixed point:} These datapoints, observed on the right bottom part of Figure \ref{fig:conv_vs_pi_analysis}, corresponds to samples where the solver nearly converges to a fixed point. This is almost always associated with high FPA scores, and good per-instance accuracy. 
    \item \textbf{Limit cycles: }These datapoints, corresponding to the right-top part of Figure \ref{fig:conv_vs_pi_analysis}, correspond to cases where the solver doesn't converge to a fixed points, but enters an orbit around it. AA scores, as well as the per-instance accuracy values remain high in this regime as well, indicating that test-time convergence (to a fixed point) is not necessary for path independence, and good accuracy. Note that the boundary between the full-convergence regime and the limit cycle regime is gradual.
    \item \textbf{Divergence:} On a number of the samples, the solver diverges. These are associated with high residuals, low AA scores and low per-instance accuracies. This shows that the main source of per-instance lack of path independence on networks that are otherwise overwhelmingly path independent (on other samples) is almost always solver divergence. 
\end{itemize}

\section{Test Time Convergence and Path independence}

\begin{figure}[!htbp]
    \centering
    \begin{minipage}[b]{0.46\textwidth}
        \centering
    \includegraphics[width=1\textwidth]{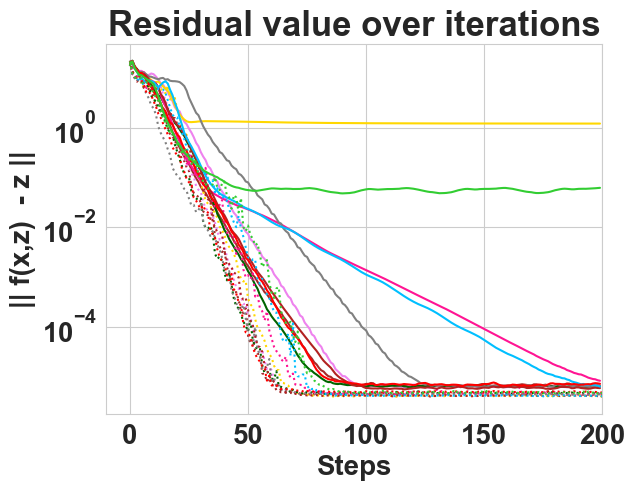}
    \subcaption[]{\label{fig:test-time-conv-res-mazes-9}}
    \end{minipage}
    \hfill
    \begin{minipage}[b]{0.46\textwidth}
    \centering
    \includegraphics[width=1\textwidth]{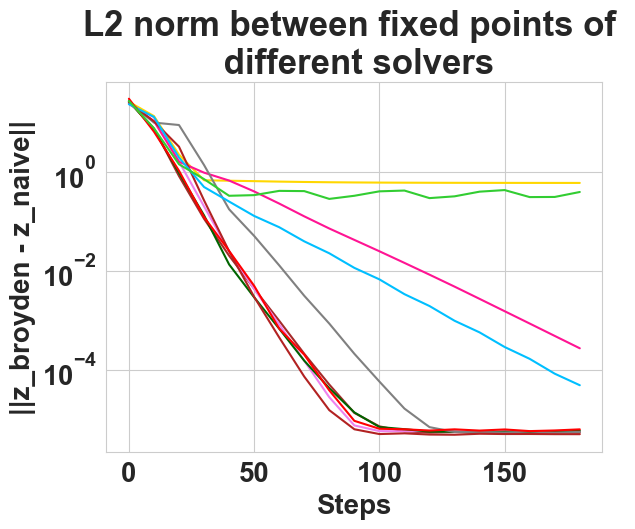}
    \subcaption[]{\label{fig:test-time-conv-euc-mazes-9}}
    \end{minipage}
        \begin{minipage}[b]{0.46\textwidth}
        \centering
    \includegraphics[width=1\textwidth]{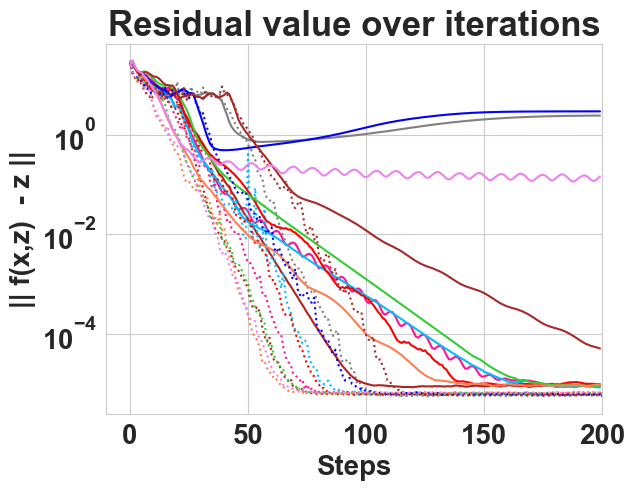}
    \subcaption[]{\label{fig:test-time-conv-res-mazes-13}}
    \end{minipage}
    \hfill
    \begin{minipage}[b]{0.46\textwidth}
    \centering
    \includegraphics[width=1\textwidth]{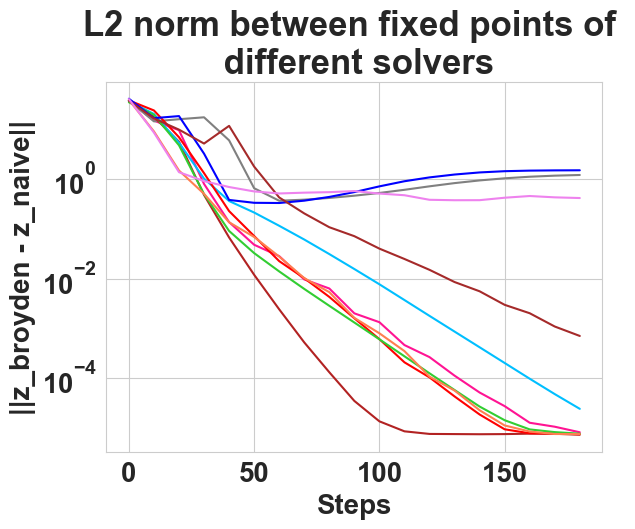}
    \subcaption[]{\label{fig:test-time-conv-euc-mazes-13}}
    \end{minipage}
    \begin{minipage}[b]{0.46\textwidth}
        \centering
    \includegraphics[width=1\textwidth]{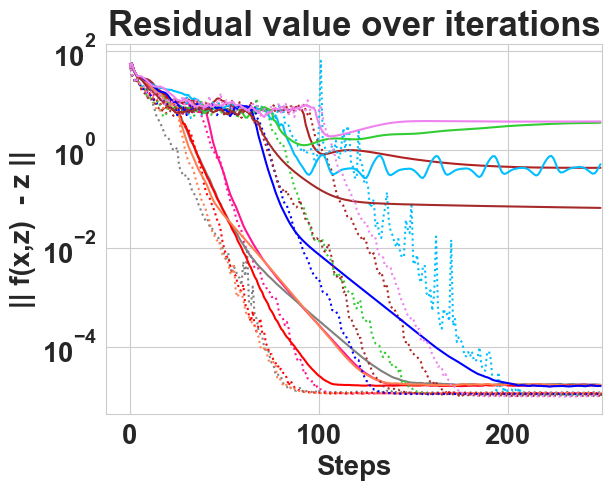}
    \subcaption[]{\label{fig:test-time-conv-res-mazes-25}}
    \end{minipage}
    \hfill
    \begin{minipage}[b]{0.46\textwidth}
    \centering
    \includegraphics[width=1\textwidth]{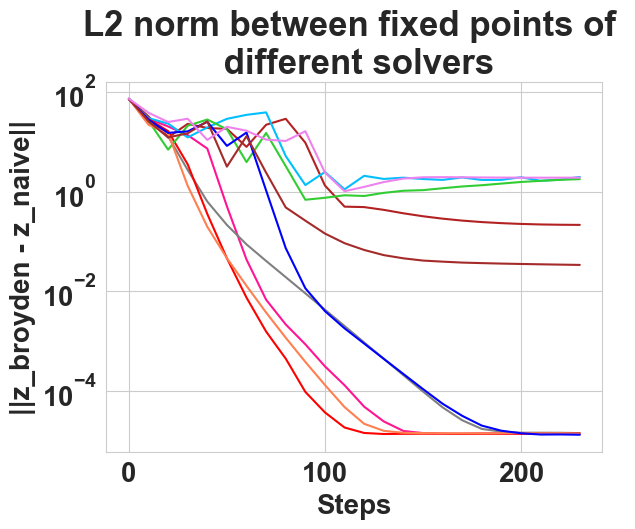}
    \subcaption[]{\label{fig:test-time-conv-euc-mazes-25}}
    \end{minipage}
    \caption{ Different solvers display differing  asymptotic behaviour but still achieve good upwards generalization (\textbf{Left column}) We display plots for the values of $||f(x, z) - z||_2$ over multiple solver steps for Broyden's method (dotted lines) and naive fixed point iterations (solid lines). (\textbf{Right column})We also plot L2 norm between the fixed points obtained through fixed point solver and Broyden's method. Each line indicates one problem instance and for a given row, lines with same color are the same problem instance. 
    The network was trained on $9\times9$ mazes, has an adversarial FPA score of 0.99, and achieves accuracy of 100\% with both the solvers on all the displayed problem instances of mazes; \textbf{(first row)} $9\times9$ mazes i.e. in-distribution, \textbf{(second row)} $13\times13$ mazes. \textbf{(third row)} $25\times25$ mazes}
    \label{fig:test-time-convergence-residuals-lr-norm-mazes}
\end{figure}

We provide a per-instance analysis of test-time convergence in Figure \ref{fig:test-time-convergence-residuals-lr-norm-mazes}. We display plots for  per-instance residuals i.e. $\| f(x,z) - z\|_2$ for Broyden's method and naive fixed point iterations over solver steps. In addition, we also plot L2 norm between the fixed points of these solvers at every step. We provide plots for both in-distribution data i.e. $9\times9$ mazes and on more difficult problem instances i.e. $13\times13$ and $25\times25$ mazes. Across all the mazes, we observe that there are problem instances where both the solvers converge to the same limiting behavior (as indicated by low values of residuals and L2 norms). However, there are a considerable number of points where naive fixed point iterations converge to a limit cycle but still output correct predictions. This behavior can be seen on both in-distribution data and on harder problem instances. We find that the absolute residuals between the points in the limit cycles are a small percentage of the Euclidean norms of the points:  0.9\% for 9x9 mazes, 0.49\% for 13x13 mazes and 1.5\% for 25x25 mazes which indicates that these limit cycles are localized. These examples of problem instances where both the solvers converge to different limiting behavior reaffirm our observation that convergence to the same fixed point at test-time is not a necessary condition for path independence.

\section{Results on the Blurry MNIST Task}
\label{app:bmnist}
We tested our path-independence hypothesis on task we call BlurryMNIST~\citep{liangout}. This task involves training an image classifier on MNIST digits corrupted with small degrees of Gaussian blur, and testing the performance on significantly more corrupted ones. The blur corruption is implemented by convolving each image with Gaussian filters of differing standard deviations. We used standard deviations from $2$ to $5.5$ (in increments of $0.5$) to generate each split. This dataset allows for testing in and out-of-distribution performance by way of training on a subset of the splits (i.e. the four lowest blur splits) and testing performance on all splits. 

\begin{figure}
  \begin{center}
 \includegraphics[width=.7\textwidth]{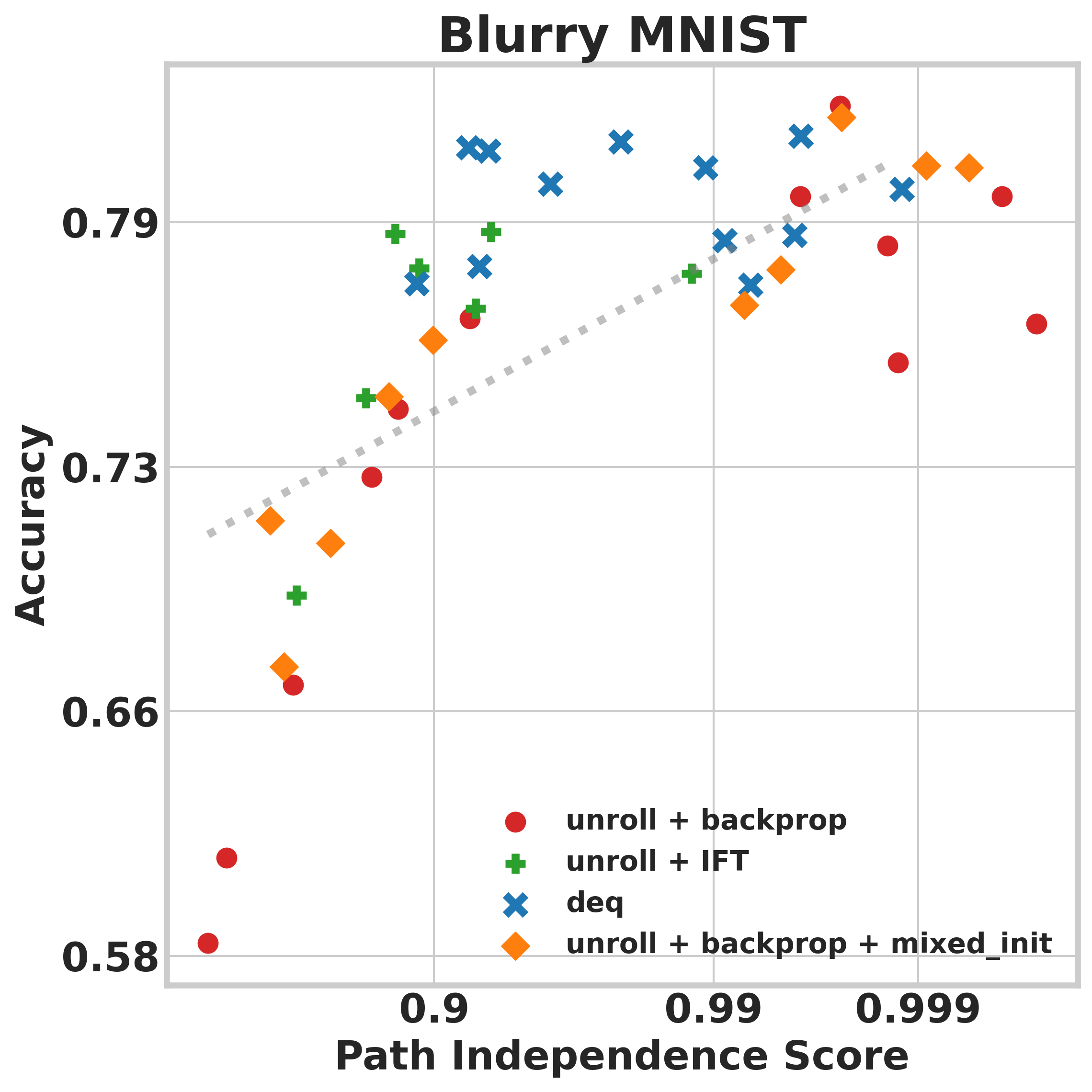}
  \end{center}
    \caption{\textbf{BlurryMNIST results:} Path independence and generalization correlate on the BlurryMNIST dataset. This task involves training an image classifier on MNIST digits corrupted with small degrees of Gaussian blur, and testing the performance on significantly more corrupted ones. } 
    \label{fig:blurry_mnist} 
\end{figure}

We trained a number of fully connected equilibrium models on the BlurryMNIST task and inspected whether path independence still correlates with accuracy. The results can be found in Figure \ref{fig:blurry_mnist}. The correlation we reported in the main body of the paper also holds in the BlurryMNIST dataset. Note that the in-distribution error rates of the trained models vary between $1$ and $5$ percent.

\section{Results on Matrix Inversion Task}
\label{app:minvert}
\begin{figure}
  \begin{center}
 \includegraphics[width=.7\textwidth]{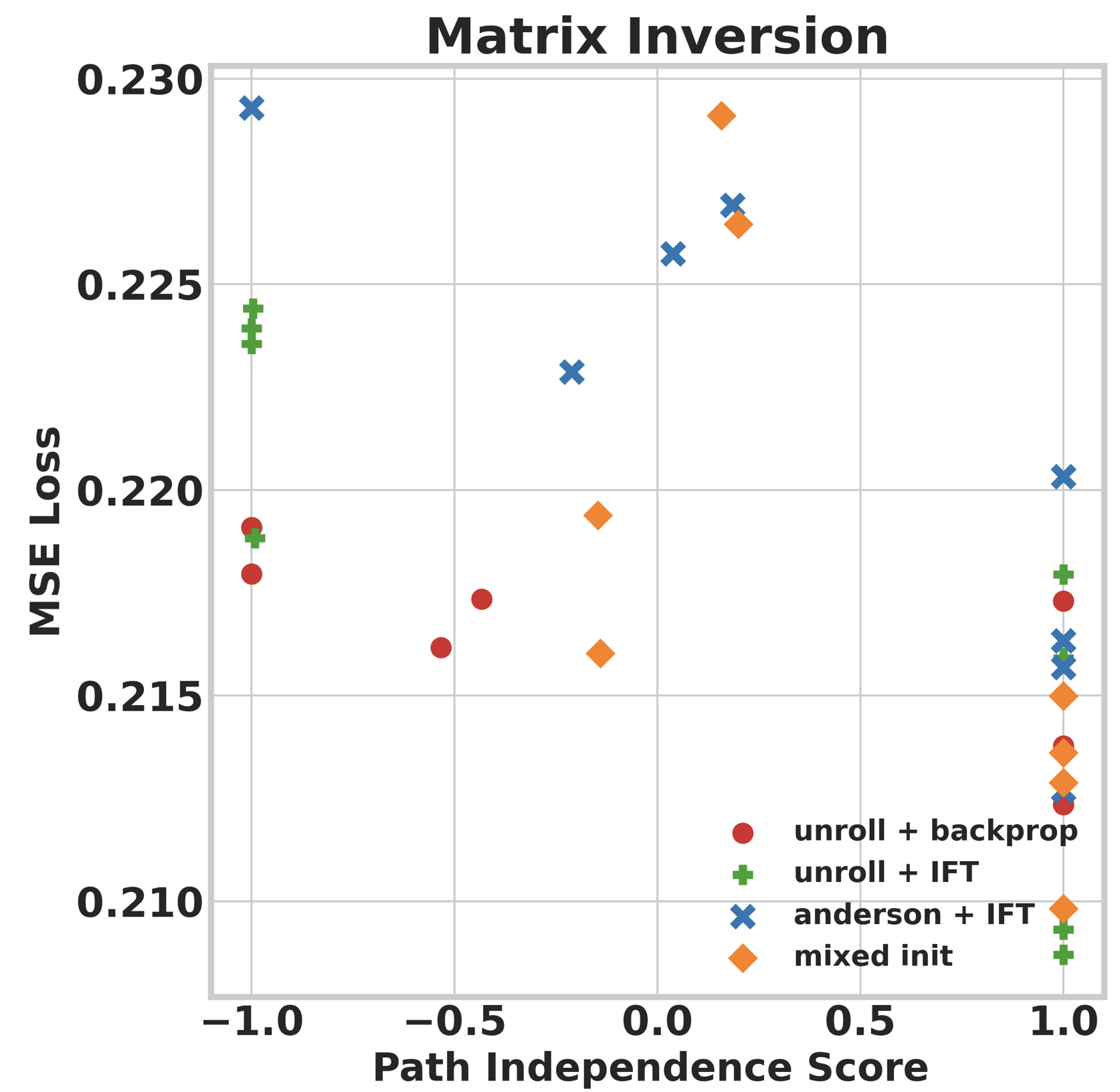}
  \end{center}
    \caption{\textbf{Matrix inversion results:} We also tested the connection between path independence and generalization on the Matrix Inversion task proposed by \citet{du2022learning}. Lack of path independence correlates with poor generalization in this task as well (note that \textbf{lower is better} in this task). } 
    \label{fig:inverse} 
\end{figure}

We also tested the connection between path independence and generalization on the Matrix Inversion task proposed by \citet{du2022learning}. This task is concerned with learning to invert 20x20 matrices. Success is defined by how well the trained model works on matrices with worse condition numbers than those observed during training. Note that this task is qualitatively very different from all the others we considered before. We have summarized the results in Figure \ref{fig:inverse}.

Using a fully-connected ResNet block as the equilibrium model cell of width $512$, we trained a number of equilibrium models where we varied 1) the forward pass (fixed point iterations vs. solver) 2) the backwards pass (backprop gradients or implicit gradients) 3) learning rate ($0.001$, $0.0001$ and $0.00001$), 4) learning rate schedule (step decays of magnitude $0.5$ at different points during training) and 5) whether layer normalization~\citep{ba2016layer}  was used or not. The results can be seen in Figure \ref{fig:inverse}. We see a similar pattern that we saw on earlier tasks: lack of path independence correlates with poor generalization. Note that our best model matches the performance of the energy based model approach proposed by \citet{du2022learning} in the matrix inversion task and significantly beats their baselines. 

\section{Results on the Edge Copy Task}
\label{app:ecopy}
We also test our path independence hypothesis on tasks that take in a graph as an input. We consider the following the edge copy task proposed by \citet{du2022learning}, where the goal is to learn to simply 
output the input edge features, in a way that generalizes to larger graph sizes.

\begin{figure}
  \begin{center}
 \includegraphics[width=1.\textwidth]{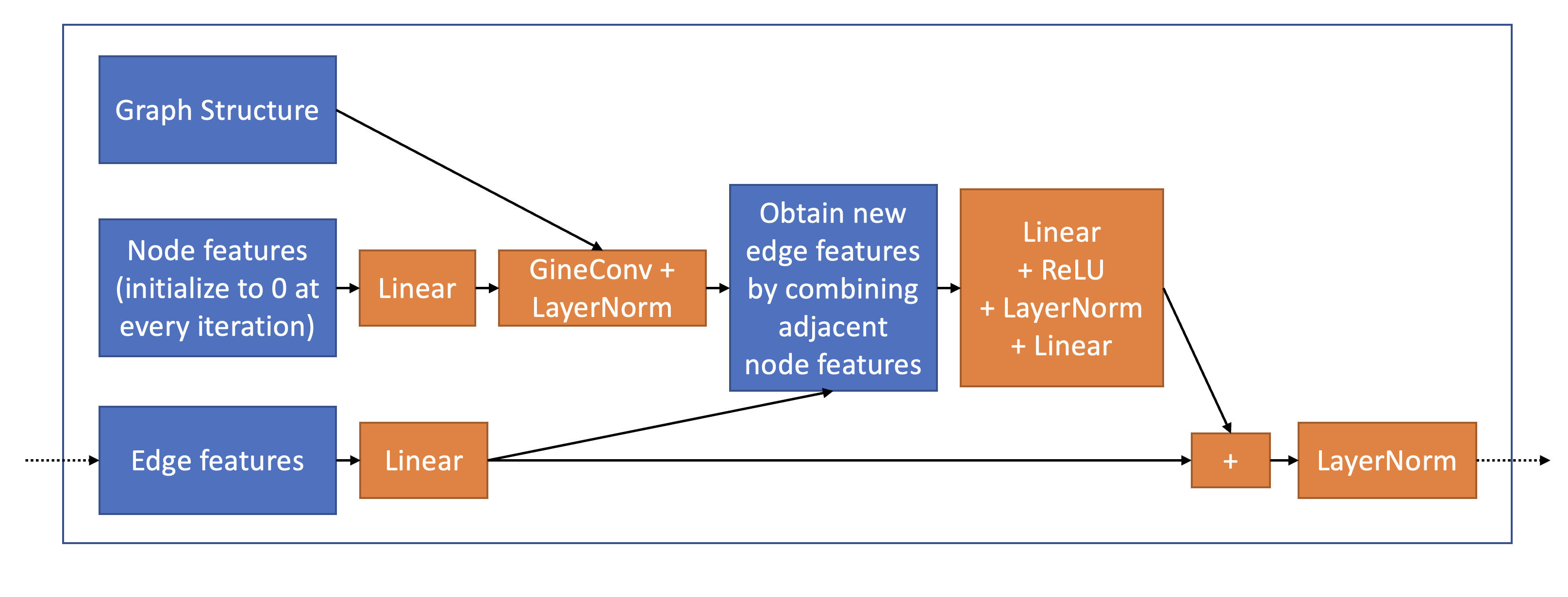}
  \end{center}
    \caption{\textbf{A graph-processing equilibrium model cell:} In the edge-copy experiments, we used the equilibrium model cell illustrated above. The workhorse of this cell is the GINEConv operation~\citep{hu2019strategies}, which fuses node and edge features to produce updated node features. This cell is especially suited for edge regression tasks, since each application of the cell refines the edge features. } 
    \label{fig:gnn_deq_cell} 
\end{figure}

We used an equilibrium model cell that's compatible with graph tasks in the Edge Copy experiments, whose structure is shown in Figure \ref{fig:gnn_deq_cell}. This cell is especially suited for edge regression tasks, since each application of the cell refines the edge features.

\begin{figure}
  \begin{center}
 \includegraphics[width=.7\textwidth]{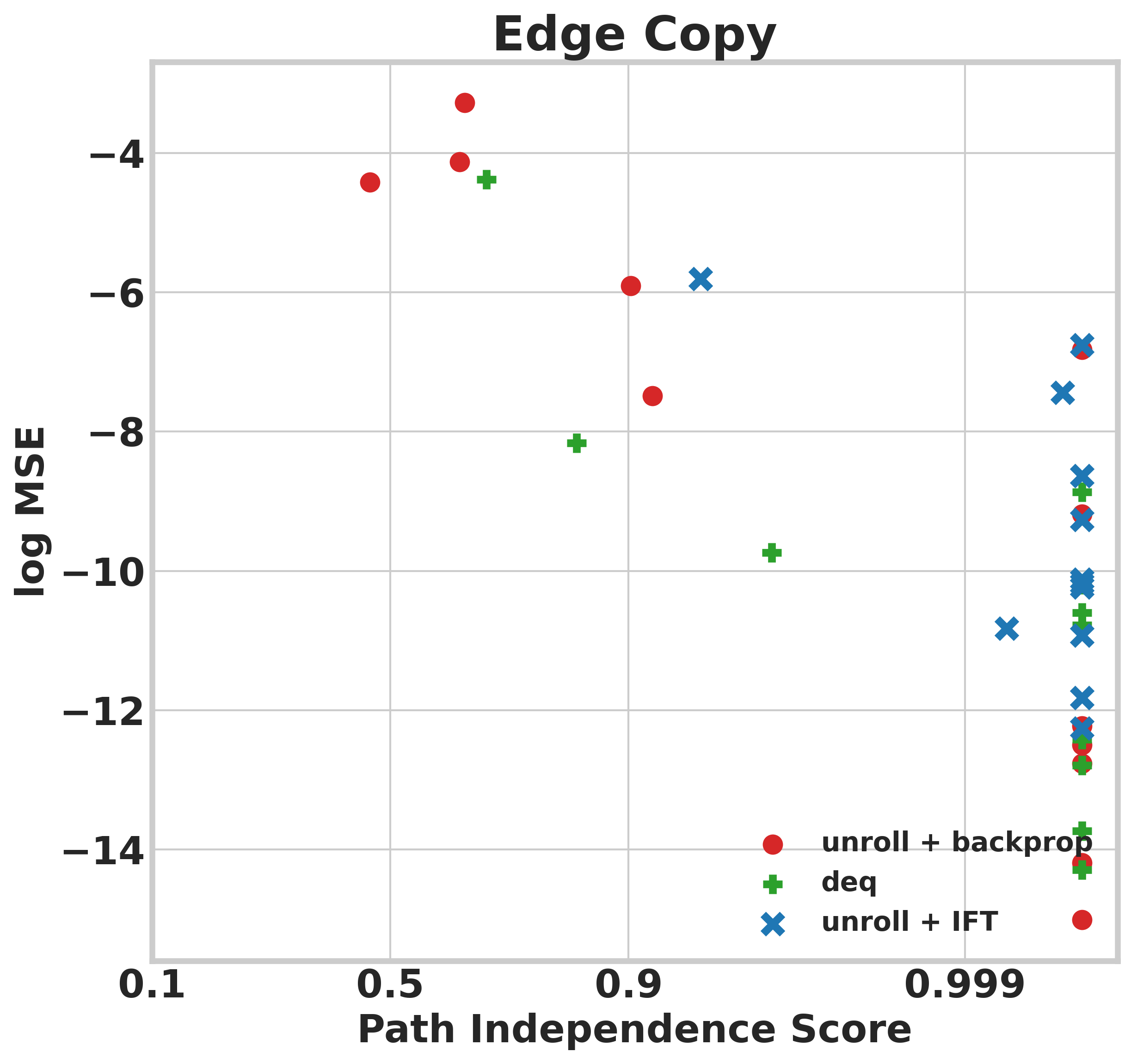}
  \end{center}
    \caption{\textbf{Edge copy task:} The edge copy task requires learning to simply 
output the input edge features, in a way that generalizes to larger graph sizes. Lack of path independence correlates with poor generalization in the edge copy task as well. Note that \textbf{lower in better} in this task. } 
    \label{fig:edge_copy} 
\end{figure}

We trained a number of equilibrium models where we varied 1) the forward pass (fixed point iterations vs. solver) 2) the backwards pass (backprop gradients or implicit gradients) 3) learning rate ($0.0001$, , $0.000333$ and $0.0001$) and 4) whether layer normalization~\citep{ba2016layer}  was used or not. Each hyperparameter configuration was run twice with different seeds. The results can be seen in Figure \ref{fig:edge_copy}. We see a similar pattern that we saw on earlier tasks: lack of path independence correlates with poor generalization. Note that our best equilibrium model outperforms the the energy based model approach proposed by \citet{du2022learning} in this task.

\end{document}